\pdfoutput=1
\documentclass[11pt]{article}
\usepackage{multirow}
\usepackage{booktabs}
\usepackage[]{acl}
\usepackage{times}
\usepackage{latexsym}
\usepackage[T1]{fontenc}
\usepackage[utf8]{inputenc}
\usepackage{microtype}
\usepackage{inconsolata}
\usepackage{graphicx}
\usepackage{xspace}
\usepackage{amsmath}
\usepackage{amsfonts}
\usepackage{booktabs}
\usepackage{multirow}
\usepackage{siunitx}

\DeclareMathOperator*{\argmin}{arg\,min}
\newcommand{\nhuman}{n_{\mathrm{human}}}
\newcommand{\E}{\mathbb{E}}
\newcommand{\thetaconfident}{\hat\theta^{\mathrm{conf}}}
\newcommand{\neff}{n_{\mathrm{effective}}}

\newcommand\blfootnote[1]{%
  \begingroup
  \renewcommand\thefootnote{}\footnote{#1}%
  \addtocounter{footnote}{-1}%
  \endgroup
}

\newcommand{\method}{\textsc{Confidence-Driven Inference}\xspace}
\usepackage{tcolorbox}
\definecolor{c1}{cmyk}{0,0.6175,0.8848,0.1490} 
\definecolor{c2}{cmyk}{0.1127,0.6690,0,0.4431} 
\definecolor{c3}{cmyk}{0.3081,0,0.7209,0.3255} 
\definecolor{c4}{cmyk}{0.6765,0.2017,0,0.0667} 
\definecolor{c5}{cmyk}{0,0.8765,0.7099,0.3647} 
\definecolor{forestgreen}{HTML}{397727}

\newtcbox{\hlprimarytab}{on line, rounded corners, box align=base, colback=c3!10,colframe=white,size=fbox,arc=3pt, before upper=\strut, top=-2pt, bottom=-4pt, left=-2pt, right=-2pt, boxrule=0pt}
\newtcbox{\hlsecondarytab}{on line, box align=base, colback=red!10,colframe=white,size=fbox,arc=3pt, before upper=\strut, top=-2pt, bottom=-4pt, left=-2pt, right=-2pt, boxrule=0pt}

\newtcbox{\hlorangetab}{on line, box align=base, colback=orange!10,colframe=white,size=fbox,arc=3pt, before upper=\strut, top=-2pt, bottom=-4pt, left=-2pt, right=-2pt, boxrule=0pt}

\newtcbox{\hlgraytab}{on line, rounded corners, box align=base,colframe=white,size=fbox,arc=3pt, before upper=\strut, top=-2pt, bottom=-4pt, left=-2pt, right=-2pt, boxrule=0pt}

\title{Can Unconfident LLM Annotations Be Used for Confident Conclusions?}

 \newcommand\coauth{$^\star$}
 \author{\hspace{0.7em} Kristina Gligorić\coauth  \hspace{0.7em}
         Tijana Zrnic\coauth \hspace{0.7em}
         Cinoo Lee\coauth   \hspace{0.7em}
         Emmanuel J. Candès \hspace{0.7em}
         Dan Jurafsky \\ 
          Stanford University\\
          \texttt{\small \{gligoric, tijana.zrnic, cinoolee, candes, jurafsky\}@stanford.edu}}

\begin{document}
\maketitle
\begin{abstract}
Large language models (LLMs) have shown high agreement with human raters across a variety of tasks, demonstrating potential to ease the challenges of human data collection. In computational social science (CSS), researchers are increasingly leveraging LLM annotations to complement slow and expensive human annotations. Still, guidelines for collecting and using LLM annotations, without compromising the validity of downstream conclusions, remain limited. We introduce \method: a method that combines LLM annotations and LLM confidence indicators to strategically select which human annotations should be collected, with the goal of producing accurate statistical estimates and provably valid confidence intervals while reducing the number of human annotations needed. Our approach comes with safeguards against LLM annotations of poor quality, guaranteeing that the conclusions will be both valid and no less accurate than if we only relied on human annotations. We demonstrate the effectiveness of \method over baselines in statistical estimation tasks across three CSS settings---text politeness, stance, and bias---reducing the needed number of human annotations by over 25\% in each. Although we use CSS settings for demonstration, \method can be used to estimate most standard quantities across a broad range of NLP problems. \end{abstract}

\blfootnote{\coauth Equal contribution.}

\section{Introduction}

Large language models (LLMs) have shown strong zero-shot performance across tasks~\cite{kojima2022large}, making them a promising tool for generating annotations, particularly when they align closely with human judgments~\cite{ziems2024can}. Given this potential, LLM annotations of textual data may be effectively leveraged for statistical estimation, hypothesis testing, and theory development~\cite{park2023generative}, as well as informing policy decisions~\cite{wei2023operationalizing}.

\begin{figure*}[ht!]
    \centering
    \includegraphics[width=\linewidth]{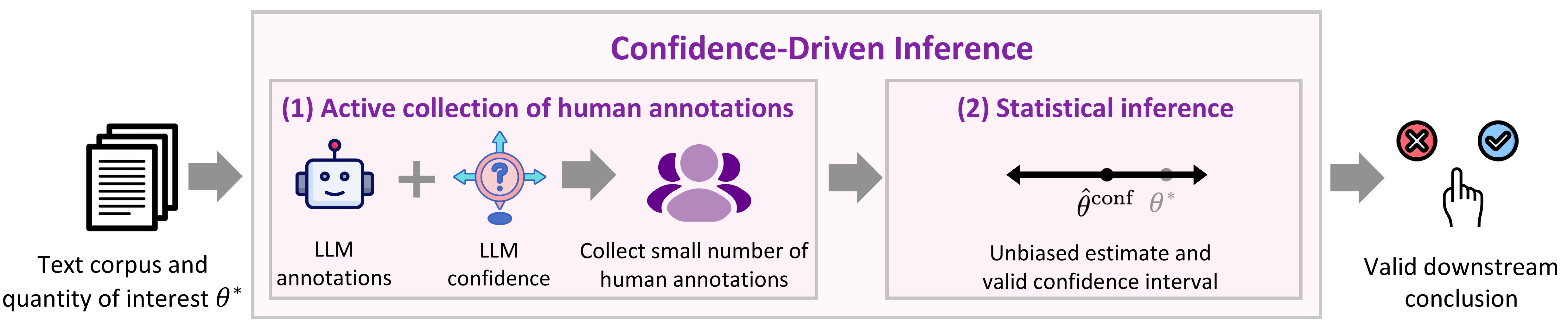}
    \caption{\textbf{Illustration of \method.} Given a text corpus and a quantity of interest $\theta^*$, (1) we collect LLM annotations and indicators of LLM confidence, based on which we strategically choose a small number of human annotations; (2) we then produce an unbiased estimate $\thetaconfident$ and a valid confidence interval, allowing valid downstream conclusions.}
    \label{fig:diagram}
\end{figure*}

Computational Social Science (CSS) research typically focuses not on the annotations themselves but on the social-science insights and conclusions they enable. Thus, understanding how LLM annotations could be used for downstream inferences is crucial in CSS. For example, stance annotations facilitate the study of linguistic differences between media affirming or denying global warming~\cite{luo2020detecting}, while politeness annotations can help examine racial disparities in verbal interactions with law enforcement~\cite{voigt2017language}, the relationship between politeness and social power~\cite{danescu2013computational}, and politeness and gender~\cite{newman2008gender}. Similarly, annotating political leanings in text allows studying the bias of search engines~\cite{robertson2018auditing}, social media~\cite{ribeiro2018media}, and political discourse~\cite{sim-etal-2013-measuring}. Precise statistical estimation, such as prevalence or regression coefficient estimation, is essential for drawing valid conclusions in such studies.
    
However, whether LLM annotations can be effectively leveraged without compromising the validity of statistical estimation remains uncertain. LLMs exhibit demographic biases~\cite{weidinger2022taxonomy,cheng2023marked} and may lack factual accuracy~\cite{gunjal2024detecting,li-etal-2023-halueval} and consistency~\cite{sclarquantifying,atreja2024prompt}. Given these limitations, using LLMs without caution may lead to inaccurate conclusions and potential societal harms, especially when such conclusions influence policy or have tangible impacts on peoples' outcomes~\cite{landers2023auditing}.
A potential solution is to rely solely on human annotations; however, human annotations are costly.


Here, we present \method, a method for valid statistical inference using LLM annotations. Given a text corpus and a quantity of interest, our approach builds on active inference~\cite{zrnic2024active} to: (1) strategically choose a small number of human annotations, guided by LLM annotations and the LLM's verbalized confidence scores, and (2) combine the human and LLM annotations into an accurate estimate of the quantity of interest (Fig.~\ref{fig:diagram}). The resulting estimate is statistically valid, while reducing reliance on expensive human annotations.

Our task is statistical estimation of a quantity of interest. We evaluate our approach on five estimation tasks in three CSS settings (politeness, stance, and media bias) in terms of confidence interval coverage and effective sample size, which measures the increase in accuracy due to augmenting human with LLM annotations (Sec.~\ref{sec:metrics}). We find that naively treating LLM annotations as human data can lead to highly inaccurate estimates and poor coverage.  By contrast, our method maintains the target coverage, while outperforming the baselines (defined in Sec.~\ref{sec:baselines}) in terms of the effective sample size. The latter is enabled partially by the fact that, in all tested settings, the verbalized confidence scores reflect LLM accuracy. Higher confidence scores correspond to higher accuracy with respect to human annotations, allowing for a strategic selection of a smaller number of human annotations.

\method can be used to estimate a wide range of standard targets (such as regression coefficients, means, and prevalences) across various NLP problems. Our code and data are available at \href{https://github.com/kristinagligoric/confidence-driven-inference}{https://github.com/kristinagligoric/confidence-driven-inference}.


\section{Background}

\subsection{LLMs for Data Annotation Tasks}

LLMs have shown great potential in handling text-annotation tasks without prior task-specific training, sometimes even outperforming crowd workers~\cite{gilardi2023chatgpt, zheng2023judging, liu-etal-2023-g,chiang-lee-2023-large, kim2023prometheus}. NLP, LLMs offer transformative opportunities for any discipline that relies on text as data. Fields such as psychology, political science, sociology, communications, and economics recognize this emerging technology's potential to enhance simulation-based research~\cite{bail2024can}, and facilitate tasks such as text analysis, concept induction~\cite{lam2024concept}, and topic modeling~\cite{pham2024topicgpt}. 

However, despite their promise, limited research has explored how to harness the potential of LLMs in ways that are both cost-effective and statistically reliable. Our work addresses this gap.

\subsection{Collaborative Annotation Paradigms}

Much of past work frames human and LLM annotations as competing alternatives, with a focus on determining which is superior~\cite{thapa2023humans}. More recent work increasingly calls for a collaborative approach that leverages the complementary strengths of both~\cite{allen1999mixed}. These collaborative paradigms aim to balance annotation quality and cost by combining human expertise and LLM efficiency~\cite{li2023coannotating, kim-etal-2024-meganno}. 

In the spirit of these collaborative paradigms, our work uses LLM confidence to efficiently and cost-effectively allocate annotation tasks, while also ensuring that the statistical inferences derived from the annotated data are valid.

\subsection{Valid Statistical Inferences in NLP}

Statistical inference is vital in NLP research. For example, model evaluation requires determining whether a model performs better than a baseline~\cite{card2020little}, which in turn relies on making valid conclusions about whether one is observing meaningful model improvements or noise~\cite{dodge2019show}. \citet{chatzi2024prediction} and \citet{boyeau2024autoeval}
leverage prediction-powered inference \cite{angelopoulos2023prediction, angelopoulos2023ppipp} for valid ranking of LLMs. A similar approach is adopted by \citet{saad2024ares} to evaluate Retrieval-Augmented Generation (RAG) systems.

Beyond model evaluation, NLP applications involve producing measurements, descriptive statistics, and causal effect estimates~\cite{feder2022causal,card2018importance}. Notably, \citet{keith-oconnor-2018-uncertainty} introduced the problem of scientifically valid prevalence estimation. They construct Bayesian confidence intervals by proposing a generative model for text documents. We contribute to the existing literature by proposing an entirely model-free approach that is applicable to a broad range of target quantities. 

Lastly, \citet{egami2024using} consider the problem of valid statistical inference when combining human and LLM annotations. However, they collect the human annotations for uniformly sampled instances, without adapting to the difficulty of annotation. Given the promise of active learning~\cite{zhang-etal-2023-llmaaa,margatina-etal-2021-active}, we develop an adaptive approach that samples a limited number of human annotations strategically. At a technical level, our approach builds on active inference \cite{zrnic2024active}, which can be seen as a refinement of prediction-powered inference \cite{angelopoulos2023prediction, angelopoulos2023ppipp} that uses active data collection for improved efficiency. Furthermore, we make use of power tuning \cite{angelopoulos2023ppipp}, a technique that ensures that incorporating LLM annotations into the estimation can never be worse than ignoring them completely.

\section{Methods}

\subsection{Problem Setup}

We have a text corpus consisting of $n$ independent and identically distributed (i.i.d.) instances $T_1,\dots,T_n$. We wish to estimate a quantity of interest $\theta^*$, such as the prevalence of political bias in the corpus or the causal effect of using certain linguistic markers on the perceived sentiment. To perform the estimation, we require human annotations $H_1,\dots,H_n$ corresponding to $T_1,\dots,T_n$. For example, $H_i$ might indicate whether $T_i$ contains political bias, or assess the perceived politeness of $T_i$. In addition to human annotations, we may also have other readily-available information about $T_i$---covariates $X_i$such as the source of $T_i$ or indicators of whether $T_i$ contains certain linguistic markers, computed via a lexicon. Note that $X_i$ is available automatically, without needing human annotation. We use the short-hand notation $T=(T_1,\dots,T_n)$ and define $X$ and $H$ similarly.

The quantity $\theta^*$ can be estimated via an estimator $\hat\theta(X,H)$, which we will denote by $\hat\theta$ for short. The accuracy of $\hat\theta$ improves as the number of samples $n$ increases ($\hat\theta$ recovers $\theta^*$ as $n$ approaches infinity). We assume that $\hat\theta$ is an \emph{M-estimator}~\cite{van2000asymptotic}, meaning it can be written as
\begin{equation}
\label{eq:full_estimator}	
\hat\theta = \argmin_\theta \frac{1}{n}\sum_{i=1}^n \ell_\theta(X_i, H_i),
\end{equation}
for a loss function $\ell_\theta$ that is convex in $\theta$. Important special cases include the mean label, $\hat\theta = \frac 1 n \sum_{i=1}^n H_i$, and linear regression coefficients, which are pervasive in CSS. Other examples include quantiles, logistic, and other regression coefficients. Notice that in some cases, like calculating the mean, the loss function only depends on~$H_i$.

Our goal is to produce an estimate of $\theta^*$ with uncertainty---by providing a confidence interval at a pre-specified level $(1-\alpha)$---with limited access to human annotations. Specifically, we can only collect $\nhuman \ll n$ annotations (on average). This means that the ``ideal estimate'' \eqref{eq:full_estimator} is out of reach.

To supplement the costly human annotations, we assume access to LLM annotations $\hat H_i$ for all $n$ instances. However, we make no assumption that the LLM annotations are good: we want to produce a valid confidence interval no matter the quality of the LLM, though we anticipate better gains when their quality is high (i.e., lower mean squared error and a smaller confidence interval).

\subsection{\method} 
\label{sec:method}

We combine LLM annotations with strategically chosen human annotations to produce an \emph{unbiased} estimate $\thetaconfident$ that lends itself to a confidence interval that is both valid and tight around~$\theta^*$. In particular, in the large-sample limit, the mean of the estimate is exactly $\theta^*$, no matter how biased the LLM annotations are.

We first explain how to choose the set of instances to be human-annotated, which is crucial for producing an accurate estimate. We collect a human annotation $H_i$ for instance $T_i$ with probability $\pi_i$. We let $\xi_i = \mathbf{1}\{H_i \text{ collected}\}$ denote the indicator of whether $T_i$ has been human-annotated.
\citet{zrnic2024active} show that the optimal choice of $\pi_i$ is to sample according to the uncertainty of the predicted annotation; roughly speaking, for most estimation problems the optimal rule is
\[\pi_i^* \propto \sqrt{\E[(\hat H_i - H_i)^2|T_i]},\]
where $\propto$ hides the normalization required to meet the budget, $\E[\sum_{i=1}^n \xi_i] = \sum_{i=1}^n \pi_i^* = \nhuman$. Of course, since $H_i$ is unknown, $\pi_i^*$ is unattainable. 

A key idea behind our method is to approximate $\pi^*_i$ by querying the LLM for \emph{verbalized confidence}. Since RLHF may cause overconfidence~\cite{geng-etal-2024-survey,zhou2024relying} and miscalibration~\cite{band2024linguistic,achiam2023gpt} of the LLM's conditional token probabilities, verbalized probabilities, i.e., expressions of confidence in token-space, are better-calibrated~\cite{tian2023just}. Therefore, to collect confidence scores, we adopt the verbalized two-stage prompting approach introduced by \citet{tian2023just}, where the model is first asked to provide an answer via zero-shooting and afterward asked to assign a probability to the correctness of the answer. This gives us a confidence score $C_i\in[0,1]$ for each instance $T_i$. In our applications, we find that the verbalized confidence scores are calibrated (Fig.~\ref{fig:confidences} (right)), meaning that higher confidence scores correspond to higher accuracy with respect to human annotations.

As we collect human annotations, we use $\{(C_j,(\hat H_j - H_j)^2)\}_{j< i, \xi_j = 1}$ as feature--label pairs to train a black-box predictor $\widehat{\texttt{err}}_i$. In other words, we train a model to predict the LLM error from its confidence. Finally, we set
\[\pi_i \propto \sqrt{\widehat{\texttt{err}}_i(C_i)},\]
normalized so that $\E[\sum_{i=1}^n \xi_i] = \sum_{i=1}^n \pi_i = \nhuman$.
In practice we do not fine-tune $\widehat{\texttt{err}}_i$ at every step $i$, but we do so periodically, after reasonably large batches of data (say, every 50 or 100 data points). See App.~\ref{app:prompts} for further details behind the sampling and Table~\ref{tab:prompts} for prompt texts.

After we have collected the human annotations according to $\pi_i$, building on active inference~\cite{zrnic2024active} we compute a \emph{confidence-driven} estimate of $\theta^*$:
\begin{equation}
\label{eq:our_estimator}	
\thetaconfident = \argmin_\theta \frac{1}{n} \sum_{i=1}^n \left(\lambda \hat \ell_{\theta,i}\mspace{-3mu}+\mspace{-3mu}(\ell_{\theta,i}\mspace{-3mu}-\mspace{-3mu}\lambda \hat \ell_{\theta,i})\frac{\xi_i}{\pi_i} \right),
\end{equation}
where we denote $\ell_{\theta,i} = \ell_\theta(X_i, H_i)$ and $\hat \ell_{\theta,i} = \ell_\theta(X_i, \hat H_i)$, and $\lambda\in[0,1]$ is a carefully chosen tuning parameter. Notice that every summand in~\eqref{eq:our_estimator} is in expectation over $\xi_i$ equal to $\ell_\theta(X_i, H_i)$, and thus the loss \eqref{eq:our_estimator} is on average equal to ``ideal'' loss~\eqref{eq:full_estimator}. This allows showing that, in the limit, $\thetaconfident$ is on average \emph{exactly} equal to $\theta^*$, no matter the bias in the LLM annotations. To give one example, if we want to estimate the mean of $H_i$, $\thetaconfident$ reduces to
\[\thetaconfident = \frac{1}{n} \sum_{i=1}^n \left(\lambda \hat H_i + (H_i - \lambda \hat H_i)\frac{\xi_i}{\pi_i} \right).\]
Notice that $\E[\thetaconfident]=\E[H_i]=\theta^*$. The parameter $\lambda$ is called a  \emph{power-tuning} parameter \cite{angelopoulos2023ppipp}, and it interpolates between ignoring the LLM annotations ($\lambda=0$) and utilizing them fully ($\lambda=1$). We set $\lambda$ \emph{optimally}, so that the mean squared error (MSE) of $\thetaconfident$ is minimized over $\lambda$. This means that, given any sampling rule $\pi_i$, the confidence-driven estimator can never be hurt by leveraging \emph{erroneous LLM annotations} or \emph{miscalibrated confidence scores}. The estimator is at least as good as when $\lambda=0$. Details behind the optimization of $\lambda$ are in App.~\ref{app:tuning}.

Finally, applying the theoretical guarantees of \citet{zrnic2024active}, we form a valid confidence interval at level $1-\alpha$ as 
\[C_{1-\alpha} = (\thetaconfident \pm z_{1-\alpha/2} \hat\sigma_{\mathrm{se}}),\]
where $z_{1-\alpha/2}$ is the $1-\alpha/2$ quantile of the standard normal distribution and $\hat\sigma_{\mathrm{se}}$ is a standard error estimate that has a closed form, stated in App.~\ref{app:confint}.

\subsection{Baselines}\label{sec:baselines}

\paragraph{Human + LLM (non-adaptive).} The first baseline incorporates LLM annotations but does not adapt to the per-instance confidence or accuracy of the LLM---it equally trusts all LLM annotations. In particular, this baseline is a special case of $\thetaconfident$ with $\lambda=1$ and uniform sampling probabilities $\pi_i = \frac{\nhuman}{n}$. This is the method evaluated and studied by \citet{egami2024using}.

\paragraph{Human only.} The second baseline ignores LLM annotations and simply applies the standard estimator to human annotations. It collects each human annotation with equal probability, $\frac{\nhuman}{n}$, so that $\nhuman$ annotations are collected on average. This is the ``classical'' approach, and it can be thought of as erring on the side of caution and ignoring potentially biased LLM outputs. Since the baseline only collects human annotations, it allows forming a valid confidence interval via classical statistics. This approach is equivalent to $\thetaconfident$ with $\lambda = 0$.

\paragraph{LLM only.} Finally, we consider the naive baseline which treats LLM annotations as human annotations, applying the standard estimator to those annotations and naively forming a confidence interval. This baseline does not suffer from a budget constraint, since LLM annotations are assumed to be cheap and available for all $n$ instances, but it may be biased.

\subsection{Evaluation Metrics}\label{sec:metrics}

We evaluate our approach and the baselines in terms of \emph{effective sample size} and \emph{coverage}. The effective sample size measures the increase in accuracy achieved by incorporating LLM annotations alongside human annotations. This is akin to getting more value out of each human annotation. For instance, if one has only 100 human annotations but combines them effectively with a larger pool of LLM annotations, the resulting accuracy could be comparable to having 150 human annotations. The latter metric, coverage, evaluates the statistical validity of the approaches by capturing how often the true value $\theta^*$ falls within the produced confidence interval. In the following we elaborate on the two metrics, deferring further details behind their computation to App.~\ref{app:metrics}.

\paragraph{Effective sample size.} Given an estimate $\hat\theta^{\mathrm{method}}$ produced by a method, we define the effective sample size as the hypothetical value $\neff$ such that $\mathrm{MSE}(\hat\theta^{\mathrm{method}}) = \mathrm{MSE}(\hat\theta^{\mathrm{human}}_{\neff})$, where $\hat\theta^{\mathrm{human}}_{\neff}$ is obtained via the human-only approach with $\neff$ annotations. In other words, $\hat\theta^{\mathrm{method}}$ is as accurate as the ``classical'' estimate with $\neff$ human annotations. An equivalent definition says that $\neff$ is the sample size for which the confidence interval around $\hat\theta^{\mathrm{method}}$ is of equal width as the classical confidence interval around $\hat\theta^{\mathrm{human}}_{\neff}$.
We thus have that $\neff-\nhuman$ is the benefit (if positive) or harm (if negative) of using LLM annotations. We also report the \emph{gain} in effective sample size, defined as $(\neff-\nhuman)/\nhuman \cdot 100\%$.
The effective sample size of the human-only approach is always $\nhuman$. We only report the effective sample size for approaches that use human annotations, i.e. all but LLM only, because the effective sample size measures the increase in value of the human annotations.

\paragraph{Coverage.} Coverage is defined as the rate at which the confidence intervals produced by each method cover $\theta^*$. Since $\theta^*$ is an ideal estimate that would require infinite data, we cannot know $\theta^*$ exactly in our applications. Instead, as a proxy, we compute coverage with respect to the estimate~\eqref{eq:full_estimator} on the full dataset. We compute the intervals with a target coverage rate of $90\%$. Note that, following the theory of \citet{zrnic2024active}, the coverage of our method is provably equal to $90\%$, and the same is true of the other two statistically valid baselines (our numbers will be slightly upward biased due to the fact that we use a proxy for $\theta^*$). With this in mind, the main purpose of reporting coverage is to evaluate the performance of the LLM only approach; for all other methods, we show coverage as a proof of concept.

\begin{figure*}[hp!]
\centering
\begin{minipage}[t]{\textwidth}
        \raggedright
\hspace{0.05\textwidth}\includegraphics[height=0.015\textwidth]{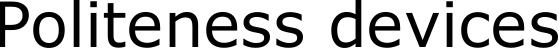}
    \end{minipage} 
\includegraphics[width = 0.3\textwidth]{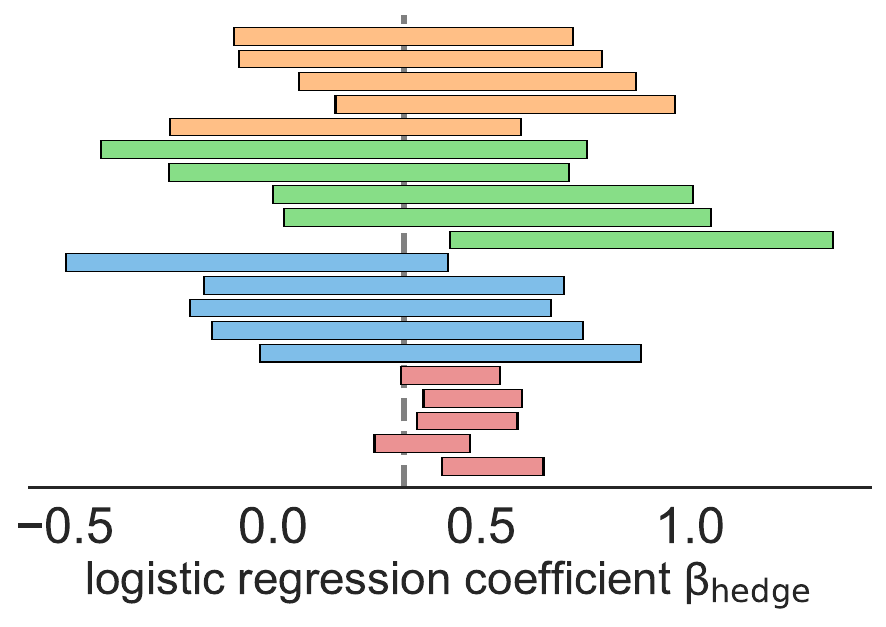}
\includegraphics[width = 0.3\textwidth]{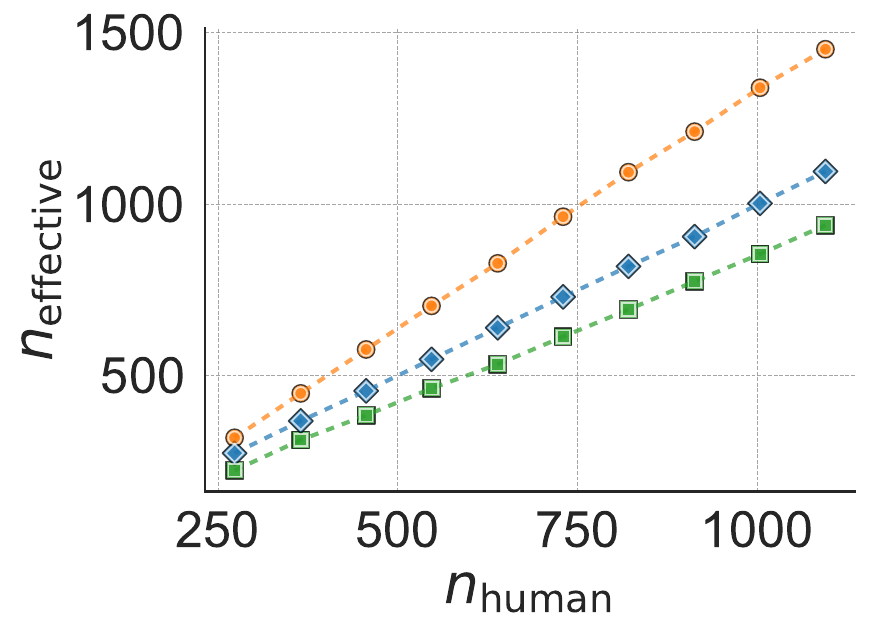}
\includegraphics[width = 0.3\textwidth]{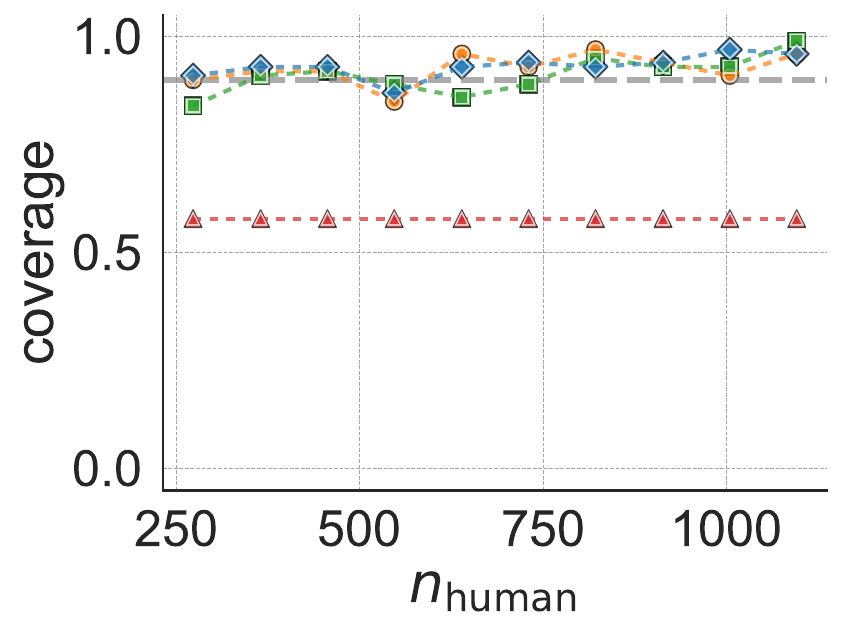}
\includegraphics[width = 0.3\textwidth]{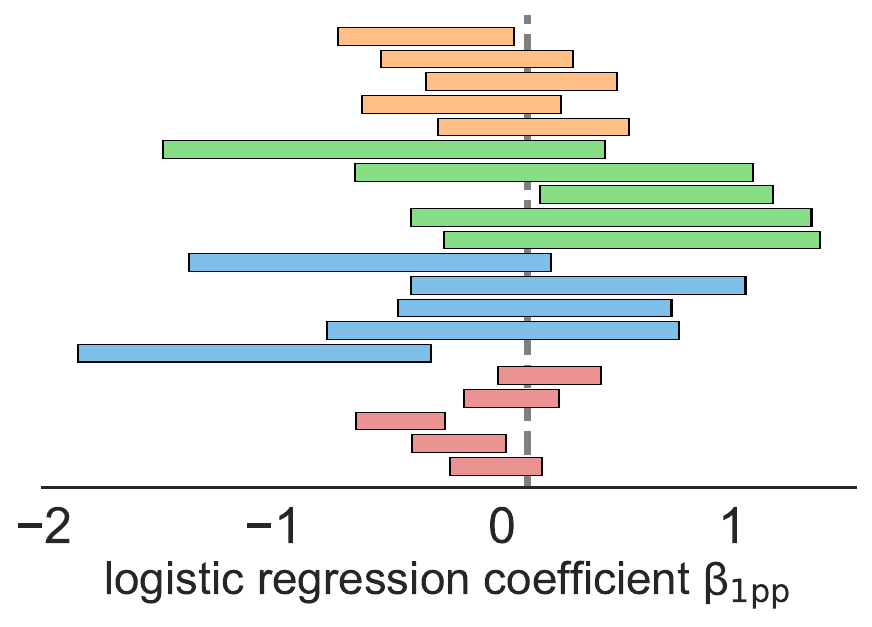}
\includegraphics[width = 0.3\textwidth]{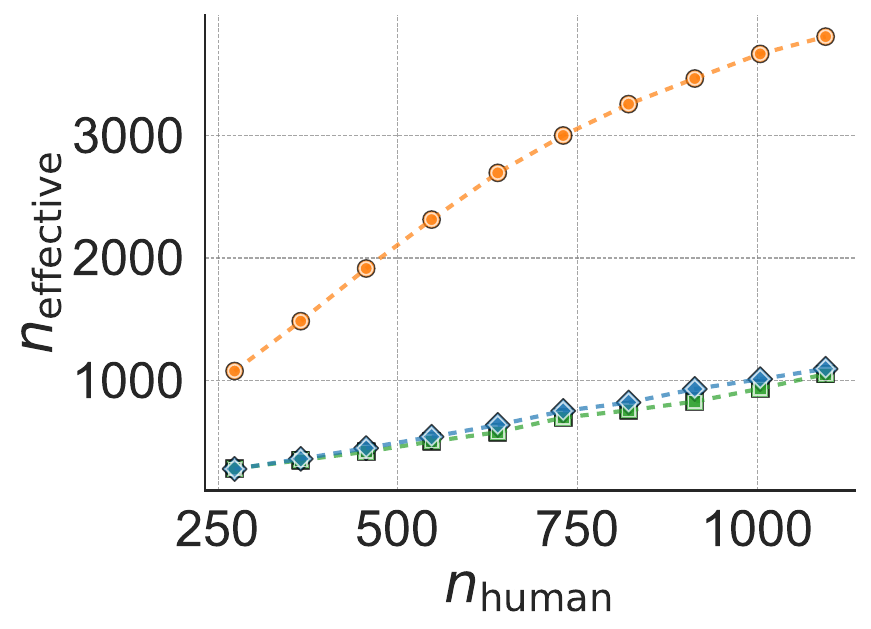}
\includegraphics[width = 0.3\textwidth]{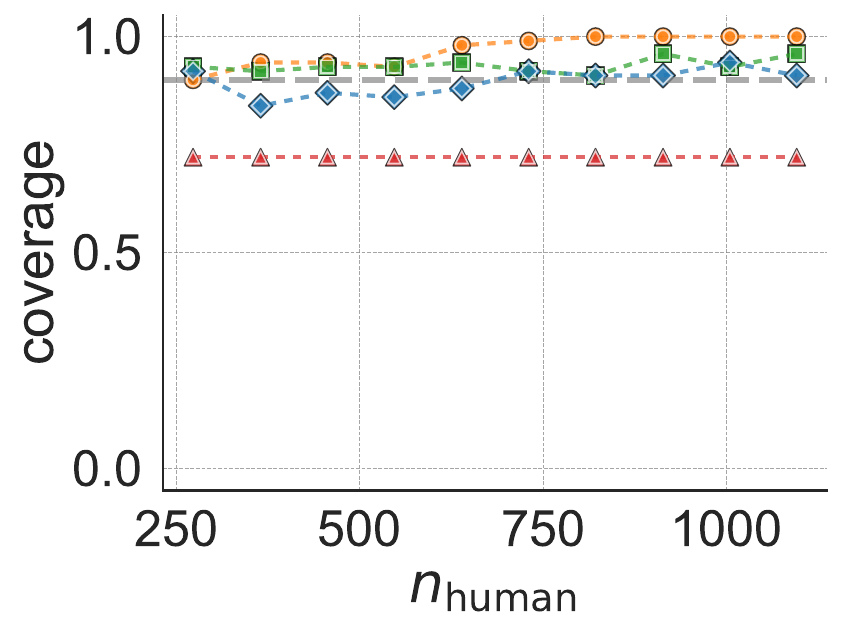}
\begin{minipage}[t]{\textwidth}
        \raggedright
\hspace{0.05\textwidth}\includegraphics[height=0.018\textwidth]{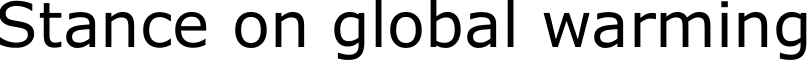}
    \end{minipage} 
\includegraphics[width = 0.3\textwidth]{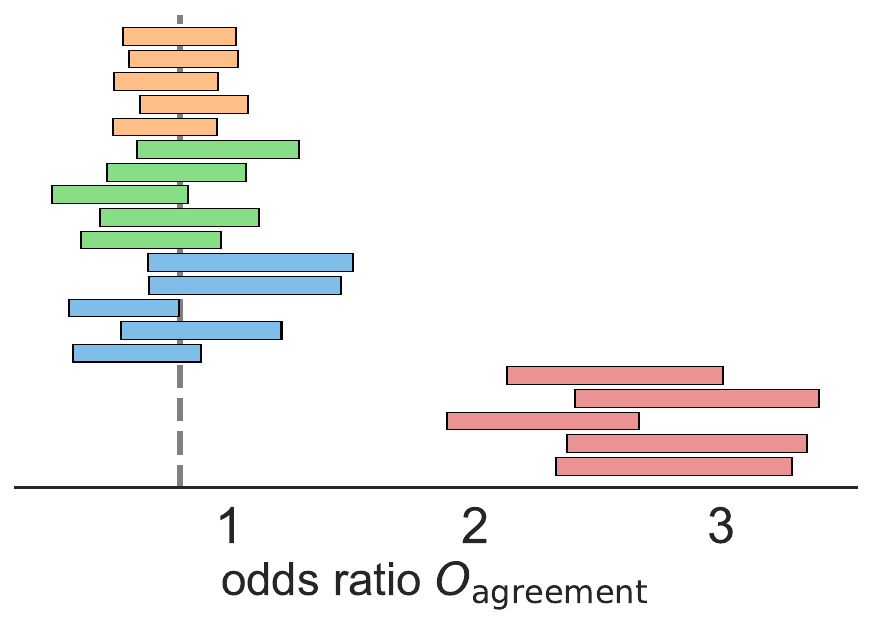}
\includegraphics[width = 0.3\textwidth]{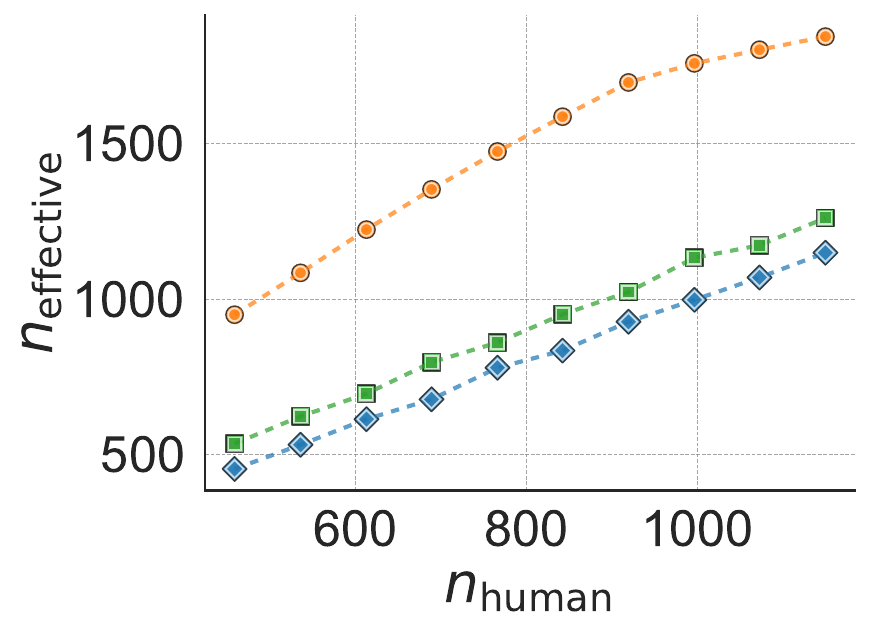}
\includegraphics[width = 0.3\textwidth]{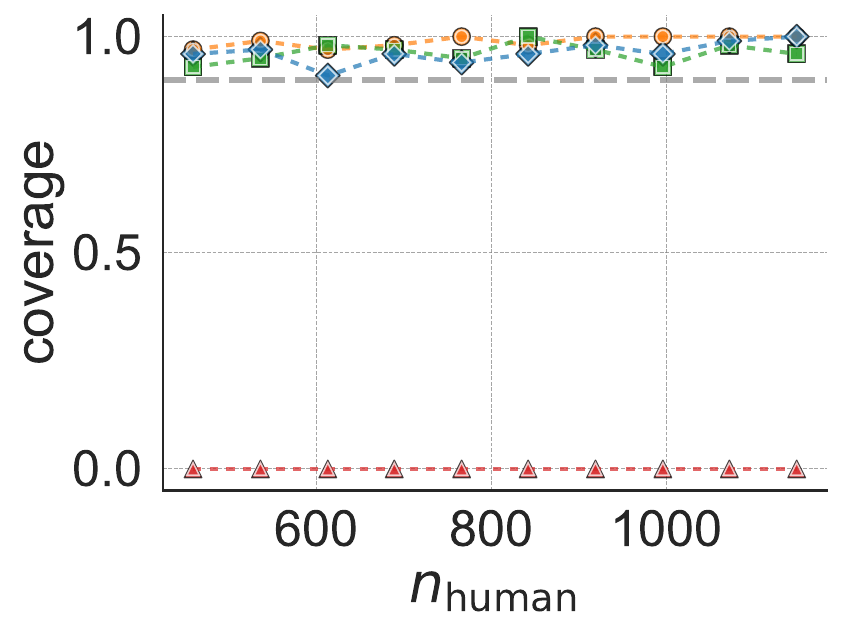}
\begin{minipage}[t]{\textwidth}
        \raggedright
\hspace{0.05\textwidth}\includegraphics[height=0.015\textwidth]{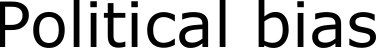}
    \end{minipage} 
\includegraphics[width = 0.3\textwidth]{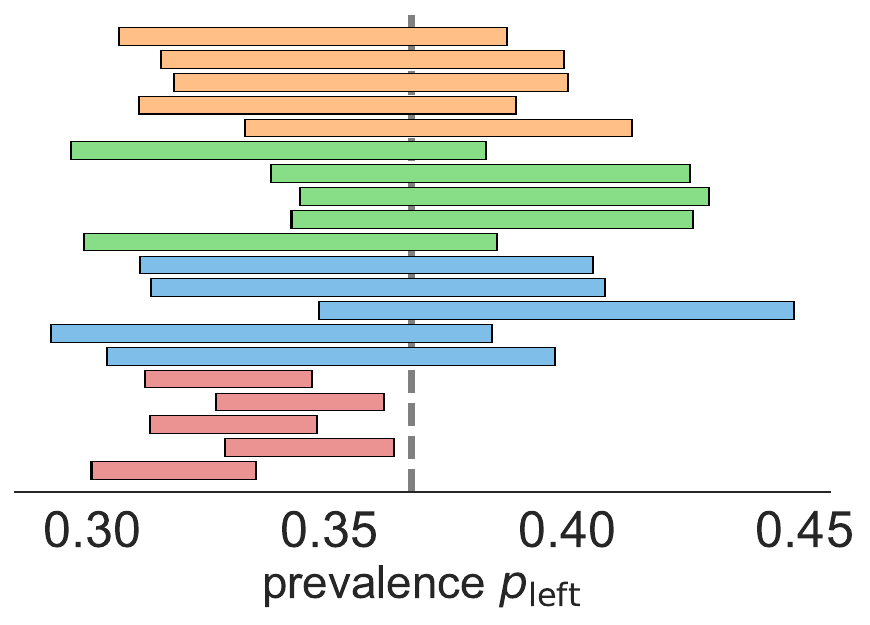}
\includegraphics[width = 0.3\textwidth]{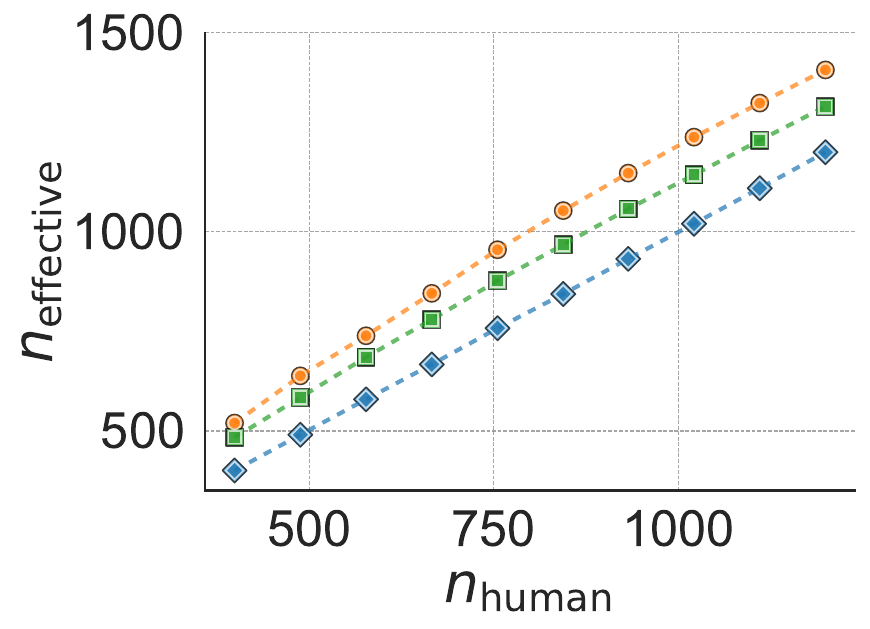}
\includegraphics[width = 0.3\textwidth]{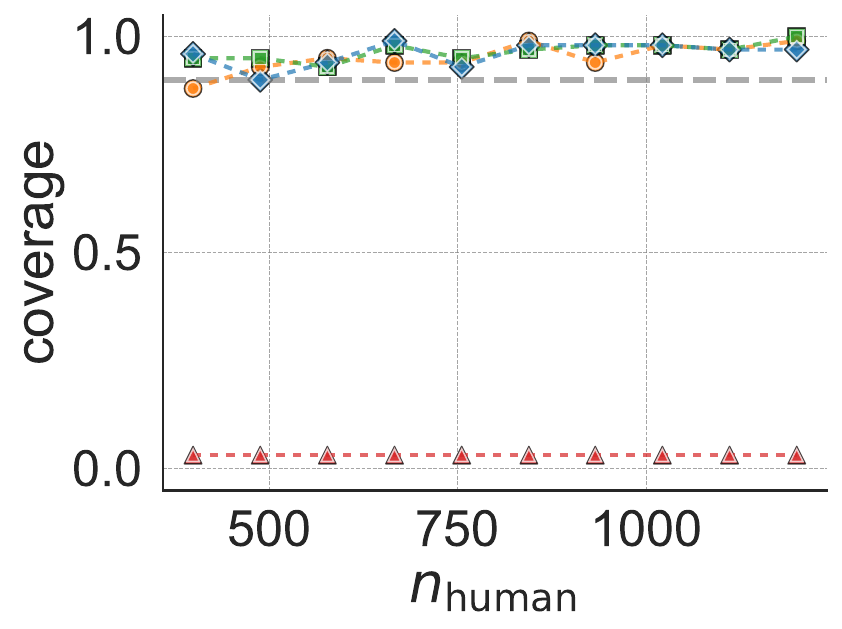}
\includegraphics[width = 0.3\textwidth]{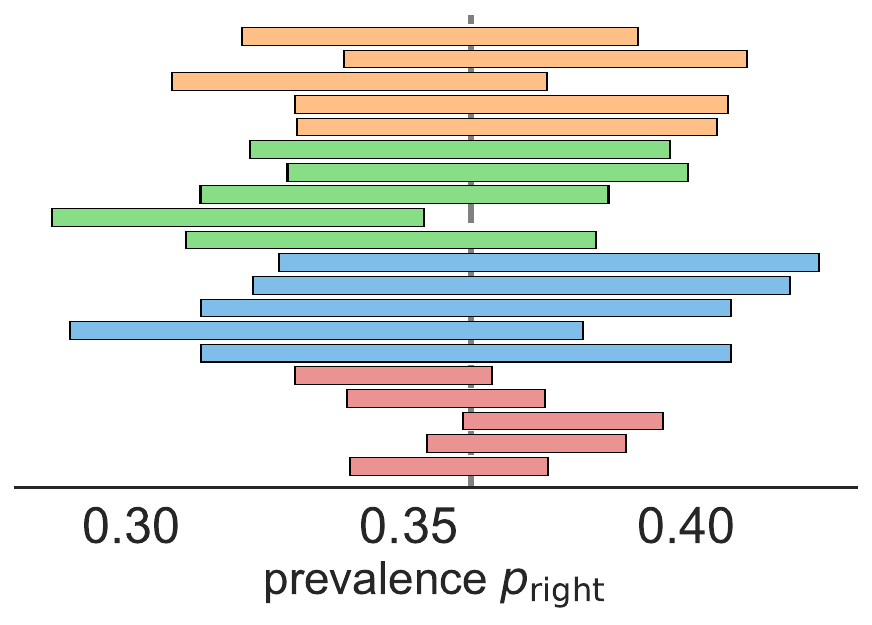}
\includegraphics[width = 0.3\textwidth]{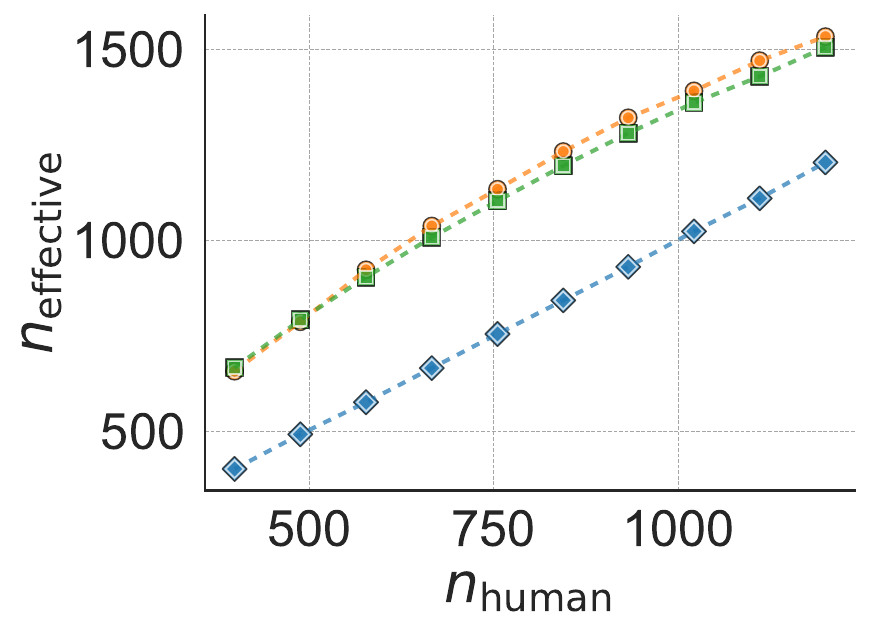}
\includegraphics[width = 0.3\textwidth]{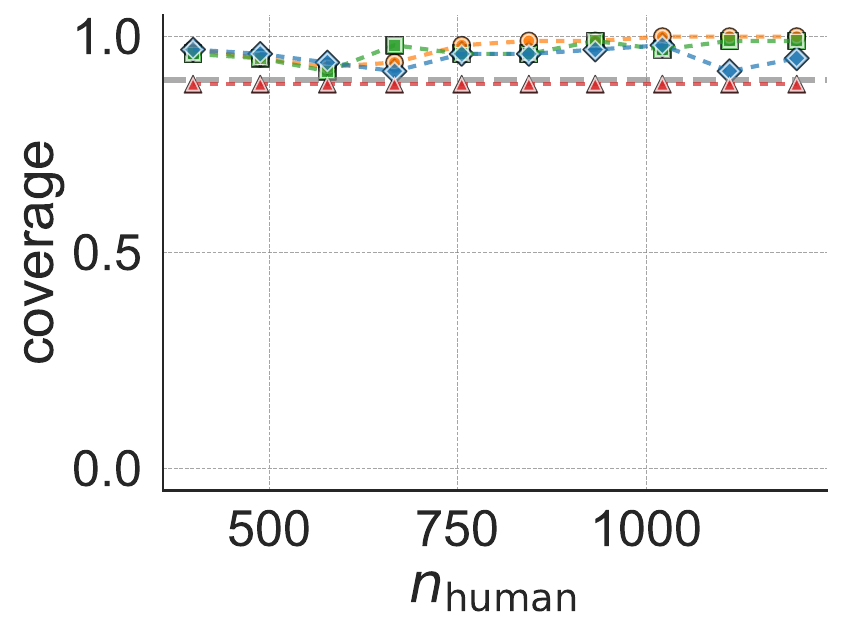}
\includegraphics[width = \textwidth]{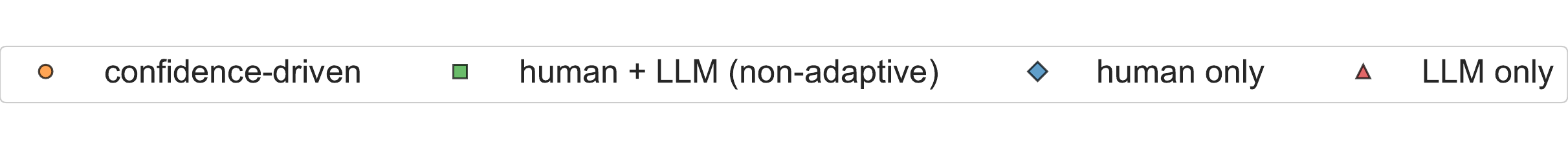}
\caption{\textbf{Confidence intervals, effective sample size, and coverage.} Rows correspond to different estimation tasks.
The first column shows the confidence intervals in five random trials. The vertical dashed line corresponds to the estimate produced on the full dataset. A method is valid if its confidence interval includes this estimate (in about 90\% of the trials), and tighter intervals around~$\theta^*$ indicates better performance. The second and third columns display the effective sample size $\neff$ and coverage, respectively, for different values of the human annotation budget $\nhuman$. Results are estimated over 100 trials.}
\label{fig:results}
\end{figure*}

\section{Results}

We evaluate our approach on a set of CSS problems that rely on statistical estimation. We aim to include settings that (1) allow addressing important downstream social-science questions, (2) rely on a human-labeled corpus of text instances (possibly with relevant additional covariates), and (3) have a publicly available dataset. We selected three settings that meet these criteria---politeness, stance, and political bias. For stance and politeness, we leverage publicly available datasets and the corresponding human annotations in their entirety. Given the large size, for political leaning, we randomly sample a smaller subset of texts. 

\subsection{Estimation tasks}

\begin{table*}[t!]
\small
\centering
\begin{tabular}{l|l|rrr}
\toprule
\multirow{2}{*}{\textbf{Estimation task}}                & \multirow{2}{*}{\textbf{Metric}}                   &     \multicolumn{3}{c}{\textbf{Method}}   \\
&  & confidence-driven & human + LLM (non-adaptive)  & LLM only \\
\midrule

\multirow{2}{*}{\textbf{\shortstack[l]{Politeness devices \\ (hedge)}}} & {Gain in eff. sample size } & (\hlprimarytab{\textbf{30.02}} $\pm$ 7.82)\%  & \negthickspace (\hlsecondarytab{-16.76} $\pm$   8.08)\%  & --- \\
& {Coverage} & \hlprimarytab{95\%} & \hlprimarytab{89\%} & \hlsecondarytab{52\%} \\
\midrule

\multirow{2}{*}{\textbf{\shortstack[l]{Politeness devices \\ (1st person pl.)}}} & {Gain in eff. sample size } & \negthinspace \negthickspace(\hlprimarytab{\textbf{319.44}} $\pm$ 22.09)\%  & (\hlsecondarytab{-8.05} $\pm$ 30.09)\%  & --- \\
& {Coverage} & \hlprimarytab{94\%} & \hlprimarytab{94}\% &\hlsecondarytab{ 69\%} \\
\midrule
                                 
\multirow{2}{*}{\textbf{\shortstack[l]{Stance on \\ global warming}}} & {Gain in eff. sample size} & (\hlprimarytab{\textbf{102.51}} $\pm$ 13.01)\% & (\hlprimarytab{13.72} $\pm$ 22.79)\%  &  --- \\
& {Coverage} & \hlprimarytab{96\%}  & \hlprimarytab{96\%} &\hlsecondarytab{0\%}   \\
\midrule
\multirow{2}{*}{\textbf{\shortstack[l]{Political bias \\ (left-leaning)}}} & {Gain in eff. sample size} & (\hlprimarytab{\textbf{29.94}} $\pm$  8.19)\%  & (\hlprimarytab{20.56} $\pm$   6.33)\%  &  ---                 \\
 & {Coverage} & \hlprimarytab{97\%} & \hlprimarytab{94\%} & \hlsecondarytab{2\% } \\
\midrule

 \multirow{2}{*}{\textbf{\shortstack[l]{Political bias \\ (right-leaning)}}} & {Gain in eff. sample size} & (\hlprimarytab{\textbf{63.73}} $\pm$ 11.48)\%  & (\hlprimarytab{61.15} $\pm$  8.75)\%  &  ---                 \\
 & {Coverage} & \hlprimarytab{91\%} & \hlprimarytab{95\%} & \hlprimarytab{90\%}  \\
\bottomrule
\end{tabular}
\caption{\textbf{Results summary.} Gain in effective sample size and coverage across the five estimation tasks for $\nhuman=500$, estimated over 100 trials.  In each task, the confidence-driven approach achieves a higher gain in effective sample size (\textbf{bolded}) than the non-adaptive approach. Confidence-driven approach always achieves a~\hlprimarytab{positive gain}, while the non-adaptive approach sometimes achieves a~\hlsecondarytab{negative gain}. Confidence-driven and non-adaptive approaches achieve \hlprimarytab{near 90\% coverage}, or higher. In contrast, LLM-only coverage is often \hlsecondarytab{poor}. Gain in effective sample size is not estimated for the LLM-only approach as it does not leverage human annotations. Errors show a standard deviation over 100 trials.}
\label{tab:summary}
\end{table*}

\paragraph{Politeness.} Texts from online requests posted on Stack Exchange and Wikipedia ($n=5,480$) can be seen as polite or impolite. Politeness annotations help understand how linguistic devices impact perceived politeness~\cite{danescu2013computational}.
In this estimation task, $\theta^*$ corresponds to the logistic regression coefficient $\beta_{\mathrm{hedge}}$ measuring the impact of a linguistic feature such as hedging on the perceived politeness, $\mathrm{logit}(P(H_{\mathrm{polite}}=~1|X_{\mathrm{hedge}})) = \beta_0 + \beta_{\mathrm{hedge}} X_{\mathrm{hedge}}$, where $X_{\mathrm{hedge}} = 1$ indicates the presence of the hedge marker and $H_{\mathrm{polite}} = 1$ indicates annotation as polite. We similarly estimate $\beta_{\mathrm{1pp}}$, the impact of the use of the first person plural pronouns on the perceived politeness.

\paragraph{Stance.} News headlines ($n=2,300$) are agreeing, neutral, or disagreeing with the stance that global warming is a serious concern~\cite{luo2020detecting}. Stance annotations facilitate the study of linguistic differences between media supporting or rejecting global warming, which have implications for communication and policy~\cite{hmielowski2014attack}. In this task, we estimate $\theta^*$ corresponding to $O_{\mathrm{agreement}}$, the odds ratio of agreement given the presence of affirming devices such as ``expert,'' ``proven,'' ``renowned,'' and so on. Formally, denoting by $X_{\mathrm{affirm}}\in\{0,1\}$ the presence of an affirming device and $H_{\mathrm{agree}}\in\{0,1\}$ the annotation of agreement, we have
\[O_{\mathrm{agreement}} = \frac{\mu_{\mathrm{agree|affirm}}/(1-\mu_{\mathrm{agree|affirm}})}{\mu_{\mathrm{agree|\neg affirm}}/(1-\mu_{\mathrm{agree|\neg affirm}})},\]
where $\mu_{\mathrm{agree|affirm}} = P(H_{\mathrm{agree}}=1|X_{\mathrm{affirm}}=~1)$ and $\mu_{\mathrm{agree|\neg affirm}} = P(H_{\mathrm{agree}}=1|X_{\mathrm{affirm}}=0)$. Indicators for affirming devices were extracted using a lexicon derived by~\citet{luo2020detecting}.

\paragraph{Political bias.} News texts (randomly sampled $n=2,000$) are either leaning left, center, or right~\cite{baly2020we}. Annotating political leanings in text allows studying the bias in media outlets, socio-technical systems, or historical and contemporary public discourse. Such biases are often reported in terms of prevalence statistics. Thus, in this setting $\theta^*$ corresponds to the prevalence of a leaning, i.e., $p_{\mathrm{lean}} = P(H_{\mathrm{lean}}=1)$, where $H_{\mathrm{lean}}\in\{0,1\}$ denotes the presence of a leaning. We estimate $p_{\mathrm{left}}$ and $p_{\mathrm{right}}$, the prevalences of left- and right-leaning articles in the corpus.

\subsection{Evaluation}\label{sec:evaluation}
Our main evaluation is based on LLM annotations collected with GPT-4o; analogous results with GPT-3.5 can be found in App.~\ref{app:models}. Table~\ref{tab:prompts} in App.~\ref{app:prompts} lists prompt texts and parameters. To test LLM performance out of the box, all annotations are collected using zero-shot prompting. Our evaluation is designed to reflect a typical CSS use-case by using standard classification CSS tasks and testing popular API-based models, without any task-specific tuning or training. Analogous results with different formulations of the annotation task, prompting mechanisms, and models are outlined in App.~\ref{sec:appendix2}.

Overall, the confidence scores are calibrated with accuracy, but the annotations are only in moderate agreement with human annotations in all three settings (see App.~\ref{app:prompts}). This is aligned with our lack of assumption that the LLM annotations are good: we want to produce a valid confidence interval no matter the quality of the LLM annotations.

We report the two key metrics (effective sample size and coverage), for the three selected settings (the study of politeness, stance, and bias), where the task is to estimate the five target quantities $\beta_{\mathrm{hedge}}$, $\beta_{\mathrm{1pp}}$, $O_{\mathrm{agreement}}$, $p_{\mathrm{left}}$, and $p_{\mathrm{right}}$.
Both metrics are estimated over 100 trials for varying $\nhuman$, the budget for human annotations. Our main findings are reported in Figure~\ref{fig:results} and summarized in Table~\ref{tab:summary}. 


\paragraph{Effective sample size.} First, across the five target quantities, we find that \method increases the effective sample size compared to the human-only baseline. For a given budget of $\nhuman$ annotations, e.g., $\nhuman=1000$, the confidence-driven approach achieves the effective sample size at minimum 1250 (when estimating $p_{\mathrm{left}}$). This means that the confidence interval around the estimated statistic is of equal width as the confidence interval produced with a larger number of human-only annotations.

Similarly, it is informative to consider the necessary budget of human annotations $\nhuman$ given a desired effective sample size $\neff$.  instance, to achieve $\neff=1000$, only between around 250 ($\beta_{\mathrm{1pp}}$) and 750 ($p_{\mathrm{left}}$) human annotations are needed. We thus reduce the number of human annotations needed to achieve equally accurate estimates by at least 25\% for all tasks.

Moreover, we also find that the confidence-driven approach increases the effective sample size compared to the human + LLM (non-adaptive) baseline. For example, to achieve $\neff=1000$, the confidence-driven approach requires 200 (respectively, 750) fewer human annotations than the non-adaptive baseline for $O_{\mathrm{agreement}}$ (respectively, $\beta_{\mathrm{1pp}}$). The confidence-driven approach therefore leads to a further reduction in the required number of human annotations compared to an approach that leverages LLMs, but does so non-adaptively. Moreover, notice that the non-adaptive approach can sometimes even hurt compared to the human-only baseline: in the two politeness tasks, using LLMs actually \emph{reduces} the effective sample size.

Table~\ref{tab:summary} summarizes the gain in effective sample size for $\nhuman=500$. Across the five tasks, the confidence-driven approach achieves a substantial gain in the effective sample size, providing at minimum around +30\% gain (when estimating $p_{\mathrm{left}}$), going even over +300\% (when estimating $\beta_{\mathrm{1pp}}$). Again, the confidence-driven approach achieves a higher gain than the non-adaptive approach for each task, which can even be negative. 


\paragraph{Coverage.} The save in human annotations does not come at the cost of diminished validity. As expected, across the five target quantities, the confidence-driven approach has coverage around or over 90\%, as do the non-adaptive and human-only baselines (Fig.~\ref{fig:results}). However, LLM-only intervals have a much lower coverage, only being around 90\% for $p_{\mathrm{right}}$, and otherwise ranging between 0\% ($O_{\mathrm{agreement}}$) and 70\% ($\beta_{\mathrm{1pp}}$). This emphasizes how estimates only relying on LLM annotations can be misleading. Notably, when estimating $O_{\mathrm{agreement}}$ using LLM annotations only, the odds-ratio estimate points in the wrong direction ($O_{\mathrm{agreement}}>1$ while $O_{\mathrm{agreement}}<1$ is true), as illustrated in Fig.~\ref{fig:results}. Interestingly, the overall inter-annotator agreement between human and LLM annotations is the highest in this setting (Cohen inter-rater agreement $\kappa_{\mathrm{stance}}=0.57$). This suggests that even when LLM annotations overall agree with human annotations, downstream statistical estimates relying on LLM annotations only can be biased. 



Table~\ref{tab:summary} summarizes the achieved coverage for $\nhuman=500$. Across the five tasks, the confidence-driven and non-adaptive approaches achieve around or over 90\% coverage (note that small deviations are possible due to only 100 simulation trials). In contrast, the LLM-only approach only meets the requirement for $p_{\mathrm{right}}$ and otherwise severely undercovers.

In summary, our method increases the effective sample size given a fixed budget of human annotations, leading to a substantial save in budget, while maintaining the target coverage. 

\section{Discussion}

In this work, we introduce \method, a method that integrates verbalized confidence of LLMs with active inference to optimally combine human and LLM annotations. Across three distinct CSS settings, results demonstrate that the proposed method consistently outperforms baseline methods (human-only and non-adaptive approaches) in effective sample size. Moreover, the increase in the effective sample size is achieved without a decrease in coverage. In contrast, the LLM-only approach yields invalid estimates and considerably lower coverage.

Wer note that the external validity of our findings is contingent upon two key assumptions: that the text instances are i.i.d. from a relevant distribution, and that the researcher has full control of the annotation process.
The first may be violated if the distribution of texts shifts over time, and the collected instances are no longer representative of the current quantity of interest. For example, it is possible that relationships between linguistic devices and perceived politeness evolve over time. The second assumption may be violated in situations where certain annotations are difficult to obtain (e.g., for low-resource languages). Our approach may lead to inaccurate or misleading conclusions under either violation. We thus caution against generalizing to settings where text instances exhibit time-varying shifts or the researcher is not in control over the data collection process.

If the adaptive sampling probabilities $\pi_i$ are poorly chosen—potentially due to inaccurate verbalized confidence scores—the resulting estimates could have a higher mean squared error (MSE) than if uniform, non-adaptive sampling were used. This could even result in an estimate with a larger MSE than the human-only baseline (for sensitivity to miscalibration, see App~\ref{app:sensitivity}). However, by using power tuning, as detailed in Section \ref{sec:method}, we ensure that incorporating LLM annotation into the estimation process does not hurt the estimate (i.e., does not increase the MSE) regardless of the sampling method used for human annotations (whether uniform or adaptive). 


Thus, \method allows for researchers to allocate human and LLM annotations in a cost-effective manner while maintaining confidence in the statistical validity of their results. Furthermore, \method also addresses the challenges posed by the variable quality of LLM annotation, by providing validity guarantees when leveraging imperfect LLM annotations. 

Although overall LLM annotations moderately agree with human annotations in the tested settings, relying on LLM annotations only can lead to wrong conclusions, as shown in the example of estimating the odds ratio in the stance setting. In contrast, despite the fact that LLM annotations are imperfect, our approach allows carefully combining them with a limited set of human annotations in order to reduce the human annotation budget, without sacrificing the validity.

Finally, the accessibility of our method is an important consideration. Across disciplines, researchers can simply prompt the LLM for its confidence via API access and leverage \method to combine LLM confidence with LLM and human annotations to produce a valid statistical estimate. This approach can be applied to a wide range of tasks, across fields.

\section*{Limitations}

We tested only a limited number of LLMs. We note that establishing a comprehensive benchmark is beyond the scope of this work (see App.~\ref{app:models} for performance details using a different model).

Additionally, while we treat human annotations as the gold standard in our study, we acknowledge that human annotations are biased, and that reasonable annotators can disagree, making it necessary to account for annotator-specific parameters~\cite{hashemi2024llm}. Future work could explore ways to account for variability and bias in human annotations. 

Human annotations are often obtained through crowdsourcing, which may itself be influenced by LLMs, as crowd workers might use LLMs to increase productivity~\cite{veselovsky2023artificial}. Although we use datasets collected before the widespread availability of LLMs, detecting AI-generated text remains a challenge~\cite{verma2024ghostbuster}. 

This work only conducted experiments on estimation tasks within CSS datasets and only in English. However, \method is generalizable to other types of text-based datasets, and it would be valuable to see more diverse applications in future research. 

Lastly, the presented experiments do not address causal effects. For instance, in the context of politeness, to identify the causal effect of hedging on perceived politeness, it would be necessary to compare texts that are otherwise identical but differ only in their use of hedging. Nevertheless, while these evaluations are not causal, our method is still applicable for use in causal estimation.

\section*{Ethical Implications}


Our work assumes that the existing human annotations within the leveraged datasets serve as the gold standard. However, we caution against interpreting human annotations as definitive judgments, given the subjective nature of many tasks~\cite{fleisig2023majority}, the potential for annotator disagreement~\cite{weerasooriya-etal-2023-disagreement}, and the influence of annotator positionality~\cite{santy-etal-2023-nlpositionality}, beliefs, biases~\cite{sap-etal-2022-annotators}, as well as variance in cultural~\cite{huang-yang-2023-culturally} and social norms~\cite{ziems2023normbank}. 

In addition to their use in text analysis, LLMs may hold potential for simulating human behavior in social science research, including applications such as pretesting surveys and imputing missing data~\cite{bail2024can}. Our work contributes to establishing reliable principles for doing so. At the same time, we do not advocate for using LLMs as substitutes for human data beyond the constraints of our assumptions, especially seeing that prior studies have shown that LLMs tend to reflect the perspectives of some demographic groups more accurately than others~\cite{santurkar2023whose} and may propagate stereotypical portrayals ~\cite{cheng2023marked}.

We also caution against fully replacing human annotators with LLM surrogates, which can not only be harmful for the economy~\cite{cazzaniga2024gen}, but also exacerbate the exploitation of human labor~\cite{li2023dimensions}. Instead, our work highlights the benefits of human-AI collaboration, showing that a combined approach can yield more accurate and valid outcomes.

\section*{Acknowledgments} This work is supported by the Swiss National Science Foundation (Grant P500PT-211127), the Stanford Institute for Human-Centered Artificial Intelligence, Stanford Data Science, and Navy Grant N00014-24-1-2305. 

\bibliography{bibliography}

\appendix

\section{Further Details on the Method}

\subsection{Confidence Intervals}
\label{app:confint}

We compute the confidence intervals following the approach in \cite{zrnic2024active}. Suppose that $\thetaconfident$ is possibly $d$-dimensional (such as in, for example, linear or logistic regression), and we are interested in coefficient $j$. If $d=1$, such as in the case of prevalence estimation, then $j$ is always equal to 1. We compute the confidence interval as:
\[C_{1-\alpha} = \left(\thetaconfident_j \pm z_{1-\alpha/2}\sqrt{\frac{\widehat\Sigma_{jj}}{n}}\right),\]
where $z_{1-\alpha/2}$ is the $1-\alpha/2$ quantile of the standard normal distribution. The matrix $\widehat\Sigma$ is an estimate of the covariance of $\thetaconfident$, given by:
\scriptsize
\[\widehat\Sigma = \hat H^{-1} \widehat{\mathrm{Var}}\left(\lambda \nabla \hat \ell_{\thetaconfident} + (\nabla  \ell_{\thetaconfident}  - \lambda \nabla \hat \ell_{\thetaconfident} )\frac{\xi}{\pi}\right)\hat H^{-1},\]
\normalsize
where $\hat H = \hat \E[\nabla^2 \ell_{\thetaconfident}]$ is the empirical estimate of the Hessian at $\thetaconfident$ and $\widehat{\mathrm{Var}}$ denotes the empirical variance. Recall also the short-hand notation $\ell_\theta = \ell_\theta(X,H)$ and $\hat\ell_\theta = \ell_\theta(X,\hat H)$.
This is a generalization of the usual ``sandwich'' covariance used in linear regression. 

Some estimation targets, such as the odds ratio, are not M-estimators but are functions of M-estimators. In those cases a confidence interval is obtained by additionally applying the delta method.

See \cite{zrnic2024active} for further details.

\subsection{Power Tuning}
\label{app:tuning}

Power tuning, introduced by \citet{angelopoulos2023ppipp}, refers to choosing $\lambda$ so that the MSE of $\thetaconfident$, or equivalently its variance, is minimized over $\lambda$. Since $\widehat\Sigma_{jj}$ is a quadratic in $\lambda$, the optimal $\lambda$ has a closed-form analytical expression. As before, suppose we are interesting in estimating coordinate $j$ of $\thetaconfident$. Let $h_j$ denote the $j$-th column of $\hat H^{-1}$. Then, we set $\lambda$ according to:
\[\lambda = \frac{h^\top~\widehat{\mathrm{Cov}}~h}{2 h^\top~\widehat{\mathrm{Var}}~h},\]
where $\widehat{\mathrm{Cov}} := \widehat{\mathrm{Cov}}(\nabla \hat\ell_{\thetaconfident} (\frac{\xi}{\pi}-1), \nabla \ell_{\thetaconfident} \frac{\xi}{\pi}) + \widehat{\mathrm{Cov}}(\nabla \ell_{\thetaconfident} \frac{\xi}{\pi}, \nabla \hat\ell_{\thetaconfident} (\frac{\xi}{\pi}-1))$ and $\widehat{\mathrm{Var}}:=\widehat{\mathrm{Var}}(\nabla \hat\ell_{\thetaconfident} (\frac{\xi}{\pi}-1))$ are empirical (co)variances. See \cite{angelopoulos2023ppipp} for further details.

\subsection{LLM and Human Annotation Details}
\label{app:prompts}

For data annotation, we use GPT-4o ({gpt-4o-2024-05-13} version) and GPT-3.5-turbo ({gpt-3.5-turbo-0125} version). Prompt texts in both stages are listed in Table~\ref{tab:prompts}. To test LLM performance out-of-the-box, all annotations are collected using zero-shot prompting. We set the max\_tokens parameter to 5, use default temperature (1), and the default system prompt and the other prompting parameters. 

Stage 1 GPT-4o annotations are in moderate agreement with human annotations in all three settings: $\kappa_{\mathrm{politeness}}=0.39$, $\kappa_{\mathrm{stance}}=0.57$, and $\kappa_{\mathrm{bias}}=0.43$.
For context, human annotators had a median inter-annotator pairwise correlation of 0.68 for the politeness dataset, while average inter-annotator agreement ranged from 0.54 to 0.64 across annotation rounds for the stance dataset\cite{danescu2013computational, luo2020detecting}. No agreement data is available for the political bias dataset.
 
\begin{figure*}[t]
    \centering
    \includegraphics[width=0.47\linewidth]{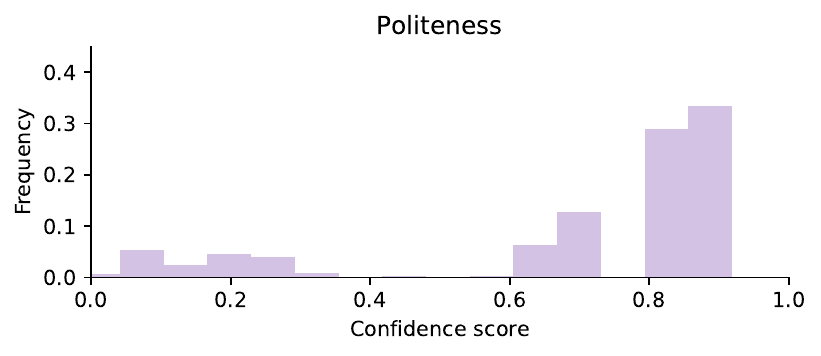}
    \includegraphics[width=0.47\linewidth]{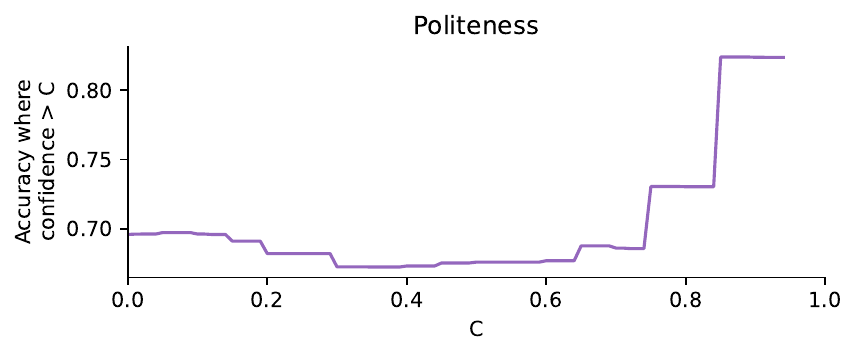}
    \includegraphics[width=0.47\linewidth]{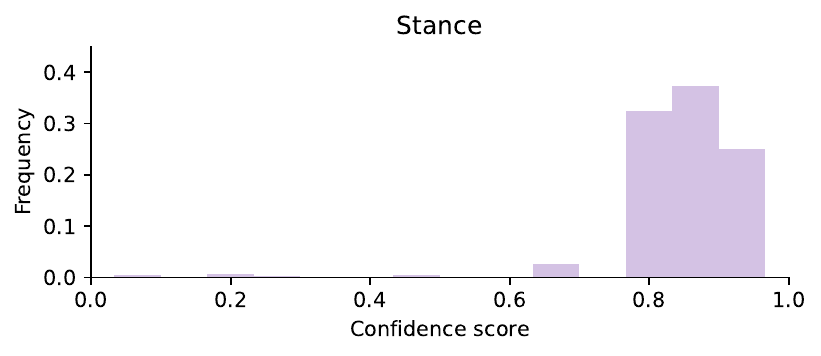}
    \includegraphics[width=0.47\linewidth]{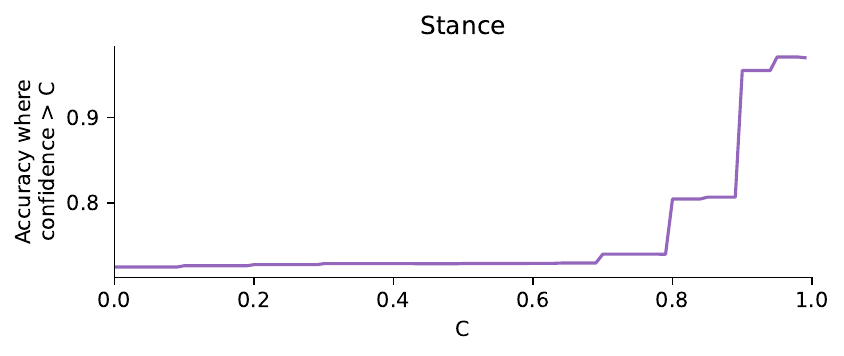}
    \includegraphics[width=0.47\linewidth]{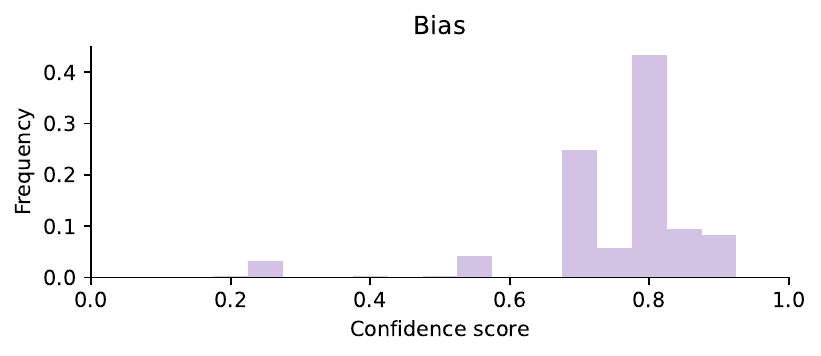}
    \includegraphics[width=0.47\linewidth]{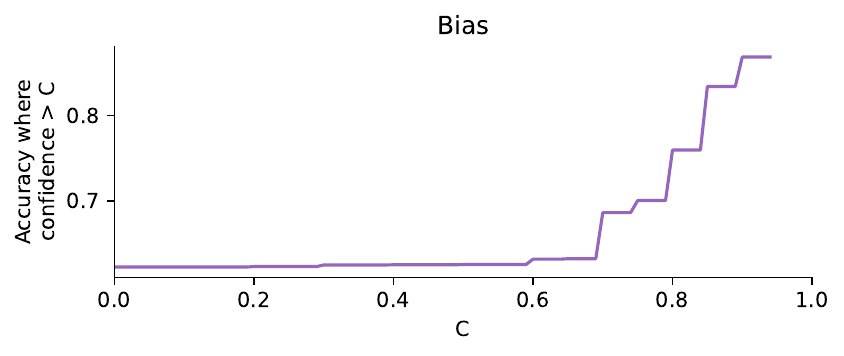}
    \caption{\textbf{Histograms and calibration curves of verbalized confidence scores.} (Left) Confidence score histograms across the three settings (GPT-4o). (Right) LLM annotation accuracy with respect to human annotations (y-axis), among instances where the confidence score is greater than C (x-axis) across the three settings (GPT-4o).}
    \label{fig:confidences}
\end{figure*}

In Stage 2, we find that the collected verbalized confidence scores are calibrated with the Stage 1 accuracy (Fig.~\ref{fig:confidences} (right)), such that higher confidence scores correspond to higher accuracy with respect to human annotations. This implies that verbalized confidence is indeed an informative signal to leverage in estimation tasks. Histograms of the collected verbalized confidence scores are illustrated in Fig.~\ref{fig:confidences} (left). We also observe a variance in the verbalized confidence within each setting, and a relative lack of overconfident responses (where the model is 100\% certain).

We choose the sampling probabilities $\pi_i$ according to the theory of \citet{zrnic2024active}. For estimating the prevalences $p_{\mathrm{left}}$ and $p_{\mathrm{right}}$, as well as the odds ratio $O_{\mathrm{agreement}}$, we choose $\pi_i \propto \sqrt{\widehat{\texttt{err}}_i(C_i)}$, as described in Section \ref{sec:method}. For the logistic regression coefficient $\beta_{\mathrm{hedge}}$ (respectively, $\beta_{\mathrm{1pp}}$), we set $\pi_i \propto \sqrt{\widehat{\texttt{err}}_i(C_i)} \cdot |X_i^\top h|$, where $h$ is the column of $\widehat H$ (defined in App.~\ref{app:confint}) corresponding to $X_{\mathrm{hedge}}$ (respectively, $X_{\mathrm{1pp}}$).

To fit $\widehat{\texttt{err}}_i$, we train an XGBoost~\cite{chen2016xgboost} model. For all problem settings, we use the same training parameters: number of boosting rounds 2000, step size 0.001, maximum depth 3, and squared-error objective.

\subsection{Computation of Evaluation Metrics}
\label{app:metrics}

We provide further details behind the computation of our two main metrics, effective sample size and coverage. For all problem settings, we run $100$ simulation trials. All experiments were run on a single CPU.

\paragraph{Effective sample size.} Recall that we define the effective sample size of a method as the hypothetical value $\neff$ such that $\mathrm{MSE}(\hat\theta^{\mathrm{method}}) = \mathrm{MSE}(\hat\theta^{\mathrm{human}}_{\neff})$, where $\hat\theta^{\mathrm{human}}_{\neff}$ is obtained via the human-only approach with $\neff$ annotations. Since all approaches but the LLM-only approach are unbiased in the large-sample limit, meaning their estimate has mean exactly equal to $\theta^*$, the MSE is simply equal to the estimator variance. Estimator variance is used in the confidence interval construction and is estimated as $\widehat{\Sigma}/n$, as explained in App.~\ref{app:confint}. The different baselines differ in their choice of $\lambda$ and $\pi$ in the definition of $\widehat{\Sigma}$. We thus compute the effective sample size as $\widehat{\Sigma}^{\mathrm{human}}_{jj}/\widehat{\Sigma}_{jj} \cdot n,$
where $j$ indexes the coordinate of $\thetaconfident$ when the estimate has more than one dimension. The final reported effective sample size is the mean of these values over $100$ trials.

\paragraph{Coverage.} We estimate coverage over $100$ trials. For all methods but LLM only, the trials differ in the random annotation decisions $\xi_i$ that determine which points get human-annotated, and we average 0/1 indicators of coverage over those trials. For LLM only, since we only have one fixed dataset of $n$ LLM annotations, in order to estimate coverage we simulate random draws from a population via the bootstrap. In other words, in each trial we draw $n$ LLM annotations with replacement, form a classical confidence interval using those points, and record a 0/1 indicator of coverage.

\section{Supplementary Results}
\label{sec:appendix2}

\subsection{LLM Data Collection Robustness}\label{app:models}

To examine the robustness of our evaluation to choices in the LLM data collection, we collected LLM annotations using varying approaches on the task of analyzing impact of hedging on perceived politeness. The default experiment (cf. Figure~\ref{fig:results}) leverages LLM annotations collected using GPT-4o model, via zero-shot prompting, where the annotation task is binary classification. In Figure~\ref{tab:robustness}, for $n_{human}=1100$, we report the gain in effective sample size and coverage using (1) an alternative smaller model (GPT-4o-mini), (2) alternative prompting mechanism (few shot prompting with ten examples), and (3) alternative annotation task (rating on a 7-point bipolar Likert scale, ranging from ``very impolite (1)'' to ``very polite (7)'').

For each LLM data collection method, the confidence-driven approach consistently achieves a higher gain in effective sample size than the non-adaptive approach. Moreover, LLM-only coverage is poor across the different data collection methods (except for the experiment with a smaller model), while non-adaptive and adaptive approaches achieve 90\% coverage or higher. We note that, although inter-annotator agreement varies substantially depending on these choices, between low ($\kappa_{\mathrm{politeness}}=0.21$) and moderate ($\kappa_{\mathrm{politeness}}=0.56$), confidence-driven approach is not harmed by the varying quality of the annotations, and always achieves a positive gain in the effective sample size. We also note that with ten few-shot examples, LLM-only coverage increases (82\%), as predicted, since examples help to guide the annotation task. We also note that LLM annotations are lower in quality when collected using a Likert scale, likely due to eliciting more fine-grained classification, which makes the rating task more challenging.

\begin{figure*}[hp!]
\centering
\begin{minipage}[t]{\textwidth}
        \raggedright
\hspace{0.05\textwidth}\includegraphics[height=0.015\textwidth]{plots/politeness-devices.pdf}
    \end{minipage} 
\includegraphics[width = 0.3\textwidth]{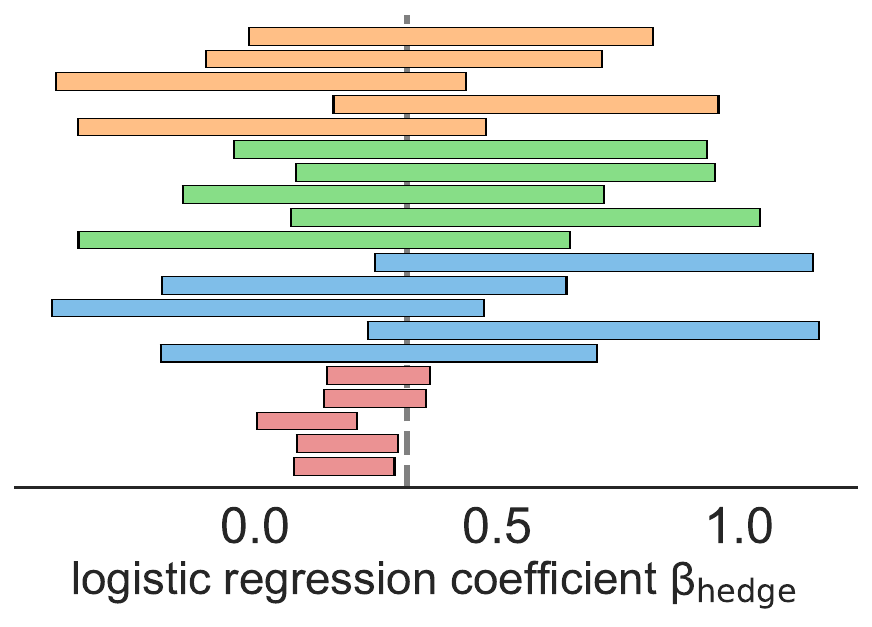}
\includegraphics[width = 0.3\textwidth]{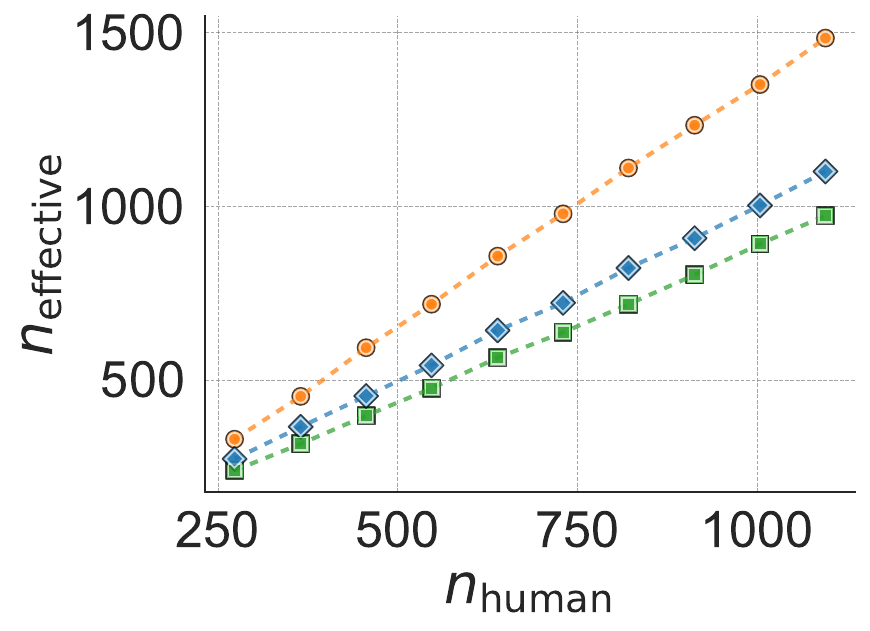}
\includegraphics[width = 0.3\textwidth]{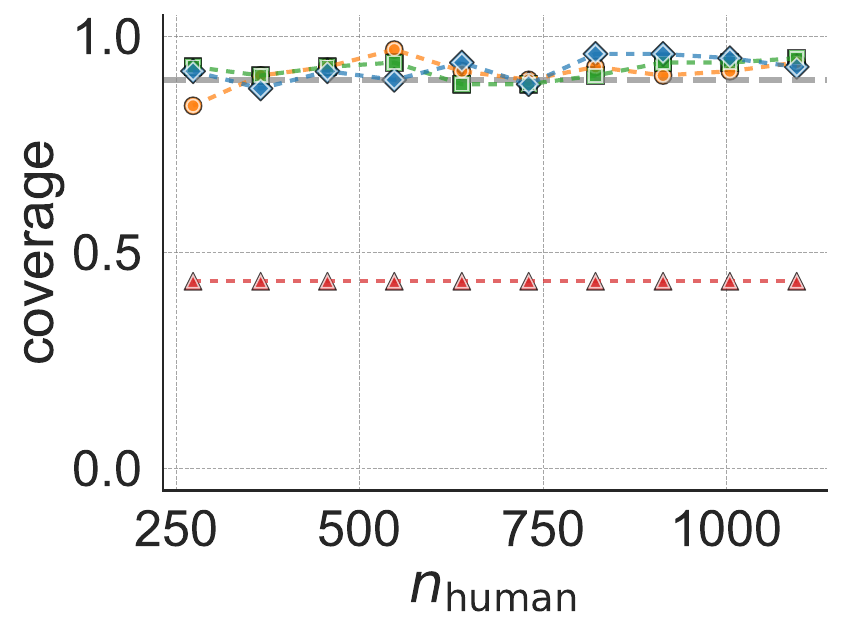}
\includegraphics[width = 0.3\textwidth]{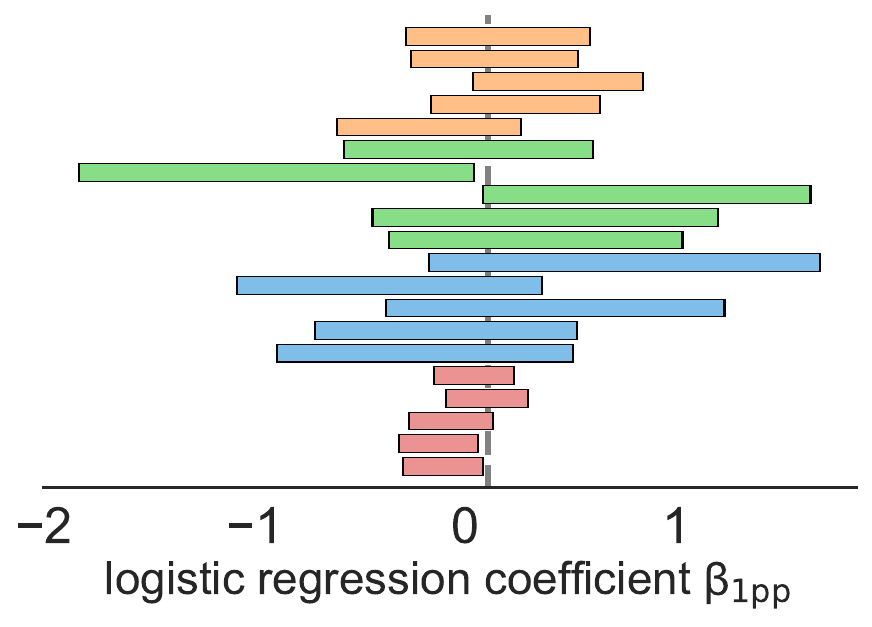}
\includegraphics[width = 0.3\textwidth]{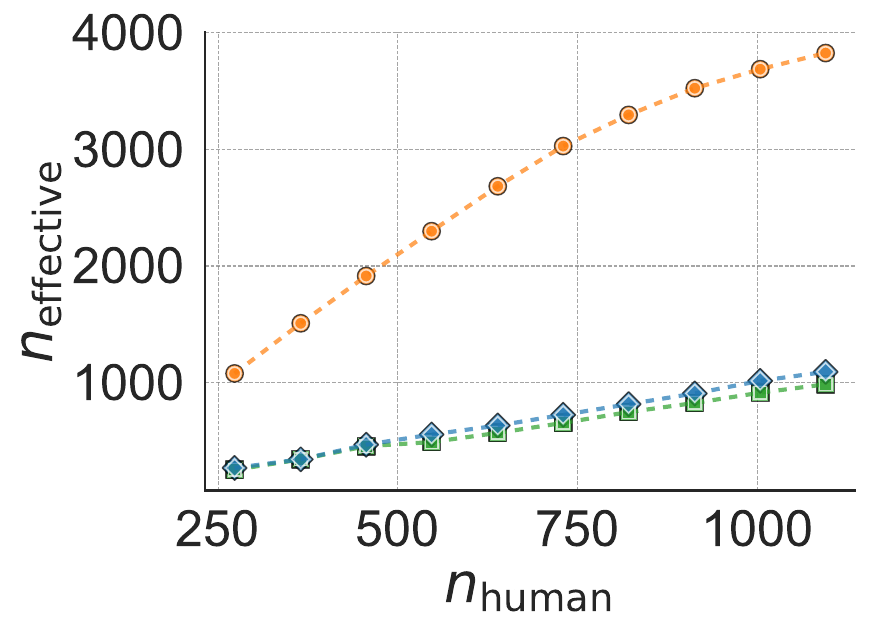}
\includegraphics[width = 0.3\textwidth]{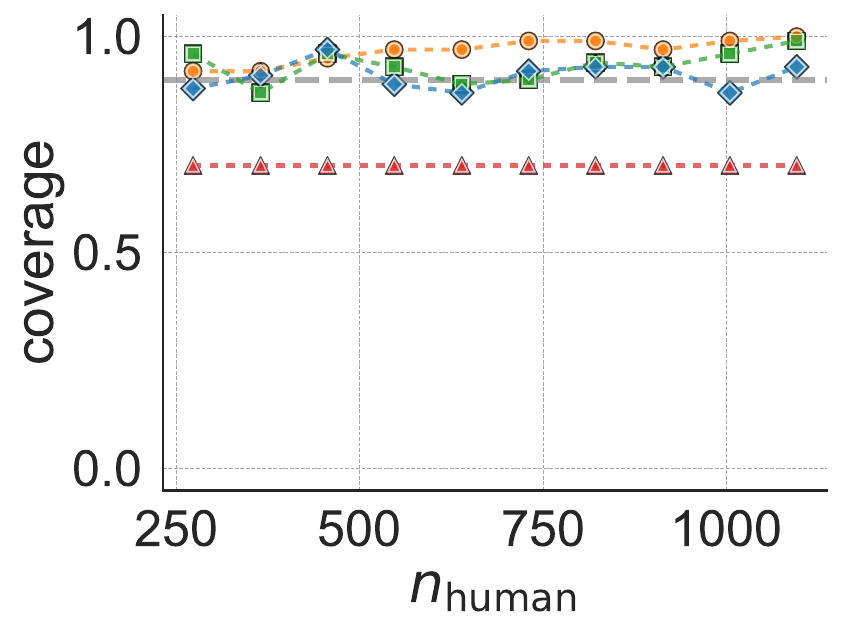}
\begin{minipage}[t]{\textwidth}
        \raggedright
\hspace{0.05\textwidth}\includegraphics[height=0.018\textwidth]{plots/stance-on-global-warming.pdf}
    \end{minipage} 
\includegraphics[width = 0.3\textwidth]{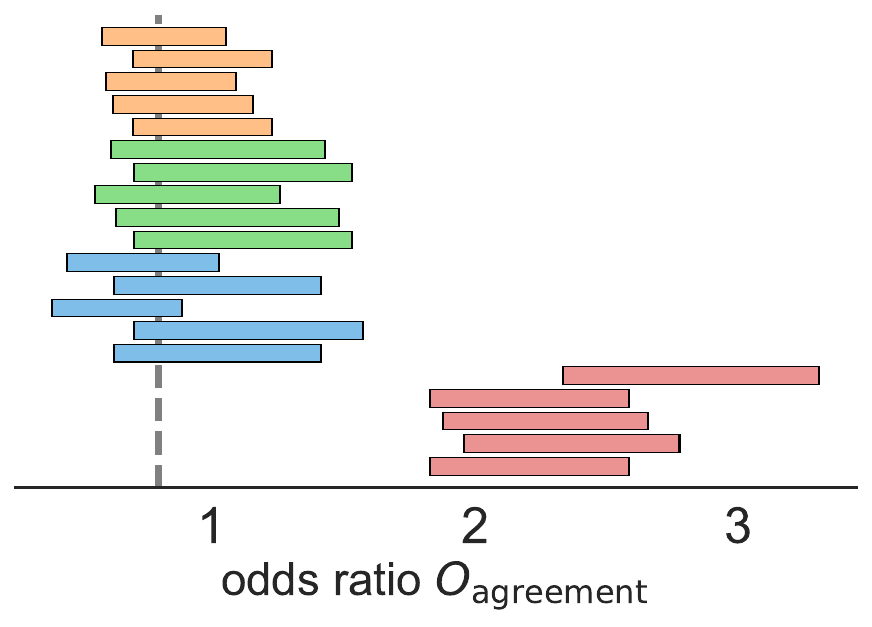}
\includegraphics[width = 0.3\textwidth]{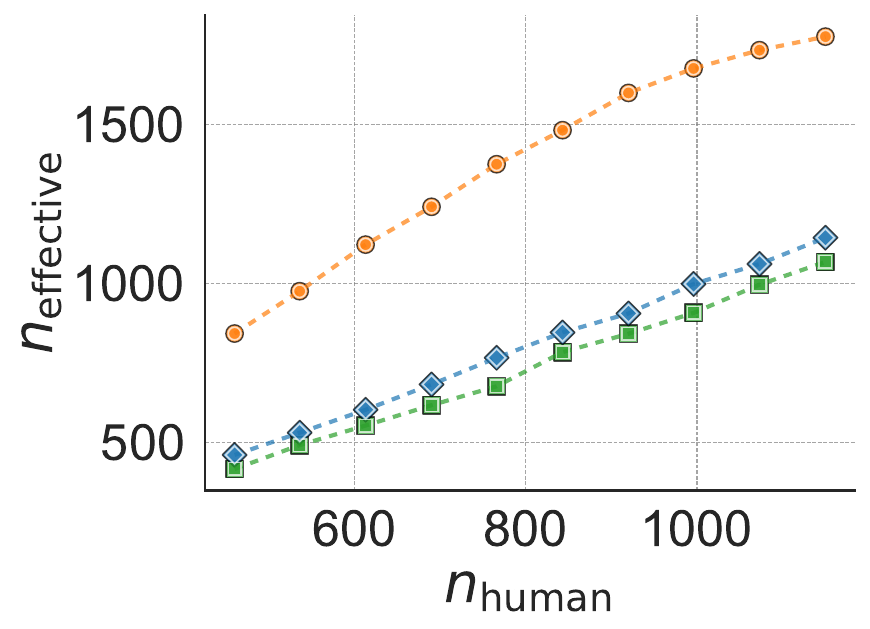}
\includegraphics[width = 0.3\textwidth]{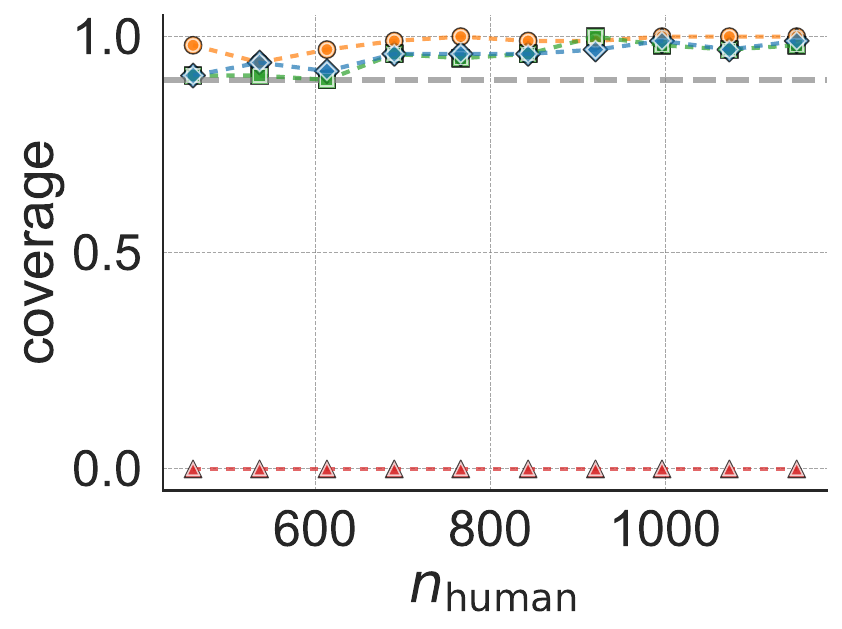}
\begin{minipage}[t]{\textwidth}
        \raggedright
\hspace{0.05\textwidth}\includegraphics[height=0.015\textwidth]{plots/political-bias.pdf}
    \end{minipage} 
\includegraphics[width = 0.3\textwidth]{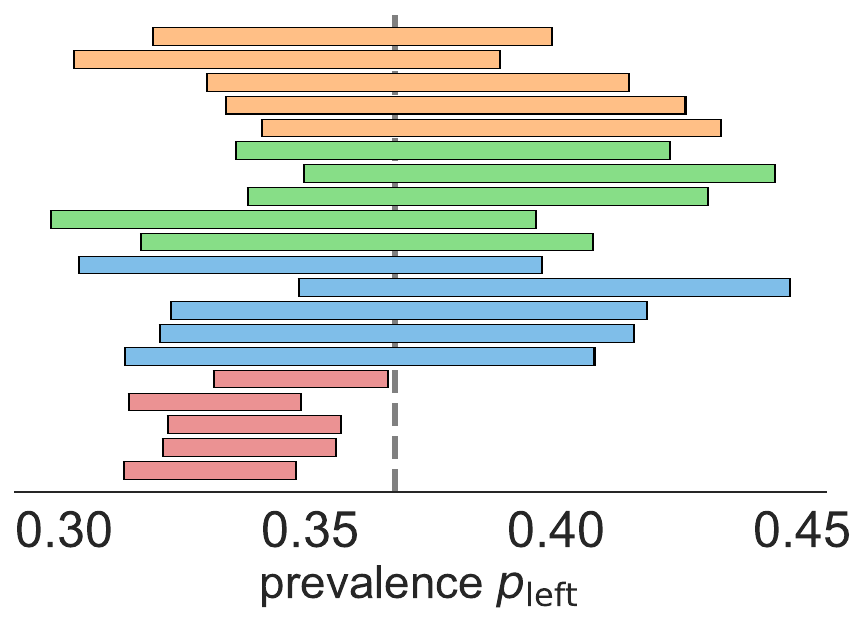}
\includegraphics[width = 0.3\textwidth]{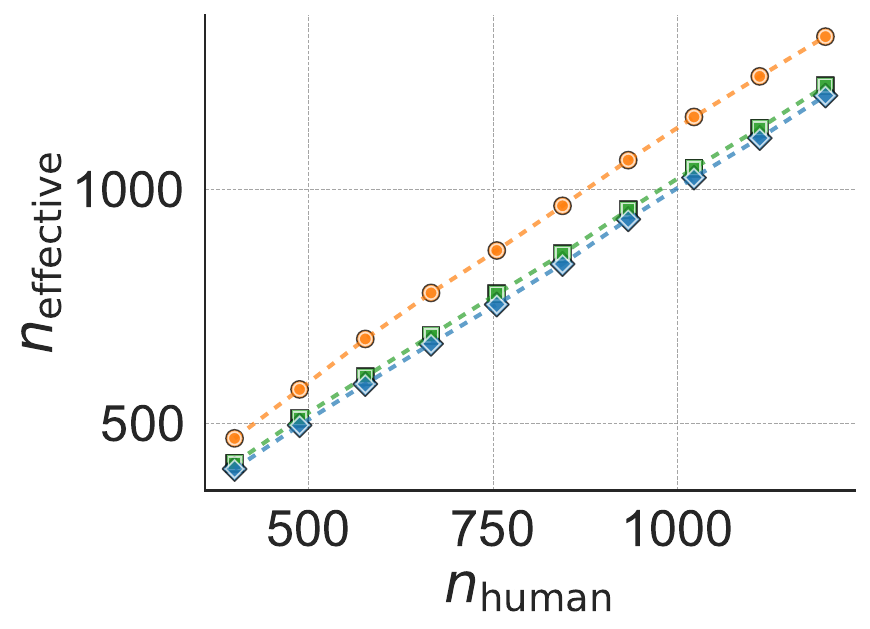}
\includegraphics[width = 0.3\textwidth]{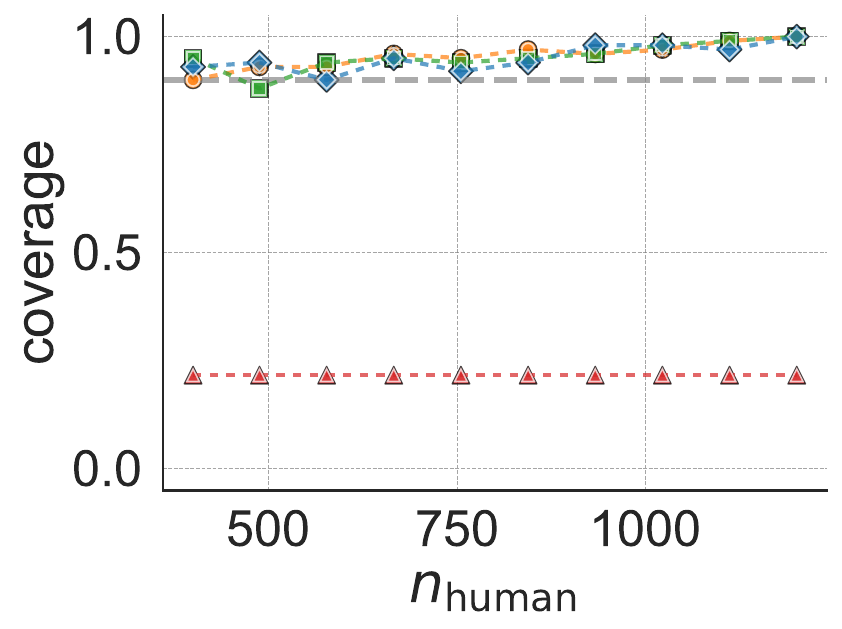}
\includegraphics[width = 0.3\textwidth]{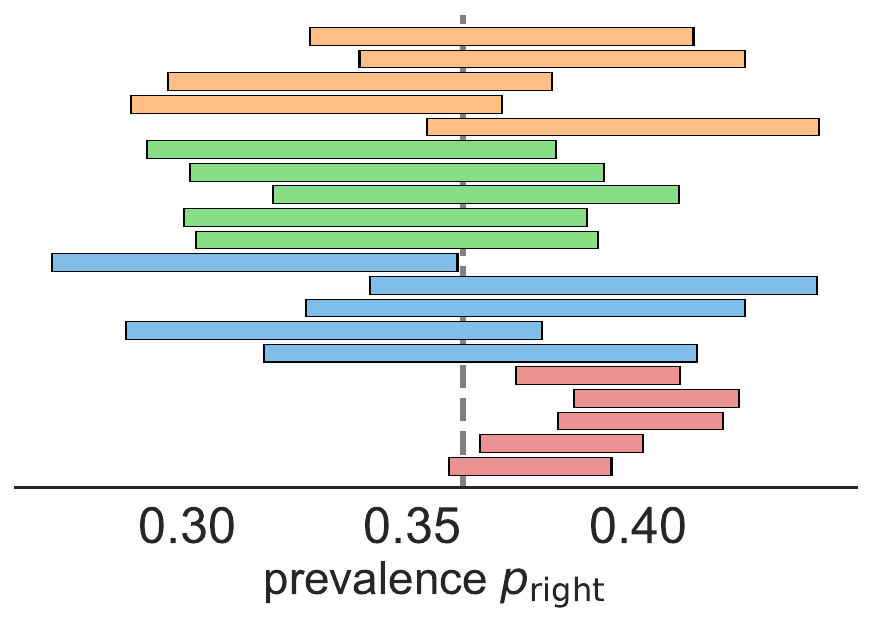}
\includegraphics[width = 0.3\textwidth]{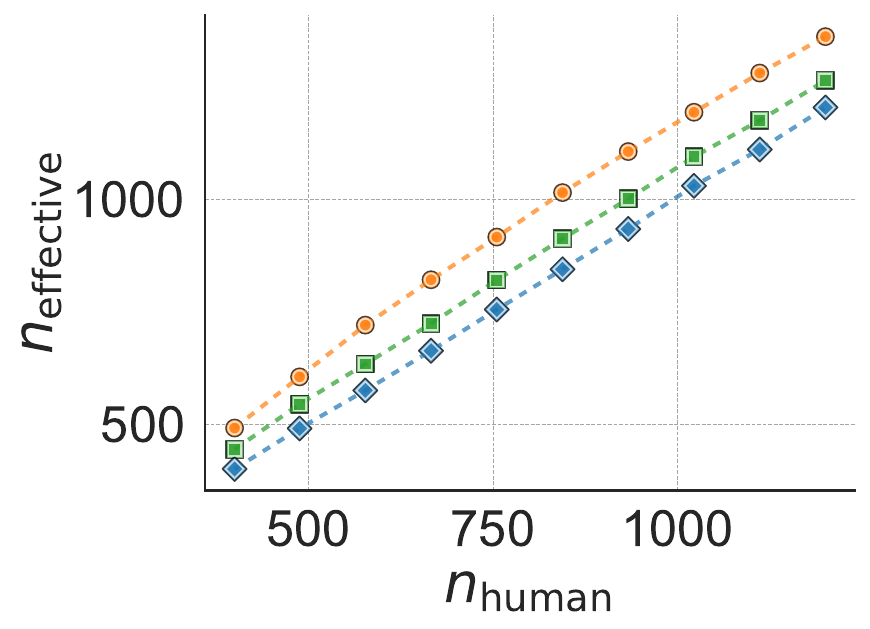}
\includegraphics[width = 0.3\textwidth]{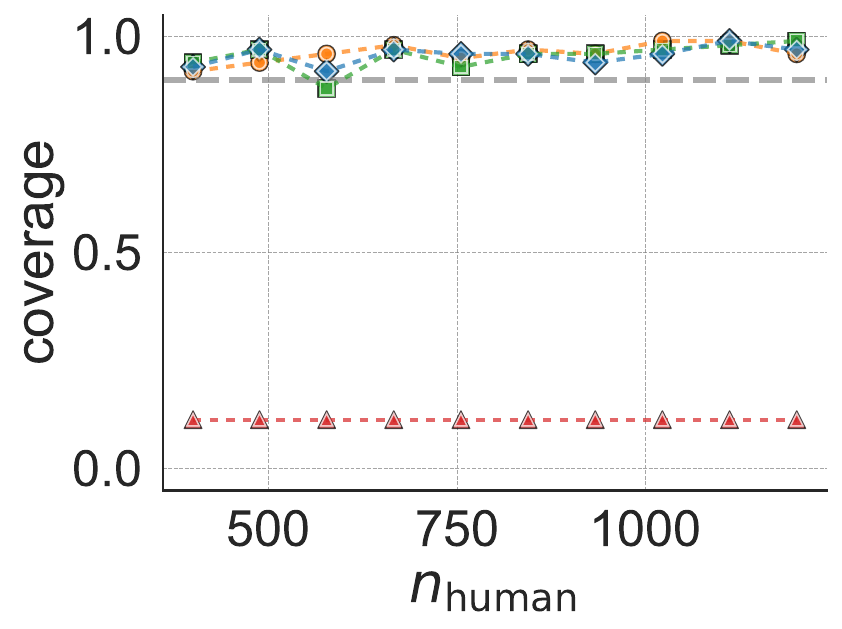}
\includegraphics[width = \textwidth]{plots/legend.pdf}
\caption{\textbf{Confidence intervals, effective sample size, and coverage (GPT-3.5).} Rows correspond to different estimation tasks.
The first column shows the confidence intervals in five random trials. The vertical dashed line corresponds to the estimate produced on the full dataset. A method is valid if its confidence interval includes this estimate (in about 90\% of the trials), and tighter intervals around~$\theta^*$ indicates better performance. The second and third columns display the effective sample size $\neff$ and coverage, respectively, for different values of the human annotation budget $\nhuman$. Results are estimated over 100 trials.}
\label{fig:3.5}
\end{figure*}

Additionally, Figure~\ref{fig:3.5} and Table~\ref{tab:3.5} summarize the results using GPT-3.5. For $\nhuman=500$, in each estimation task the confidence-driven approach again achieves a higher gain in effective sample size than the non-adaptive approach. Moreover, it always achieves a positive gain. In contrast, the non-adaptive approach achieves a negative gain in three out of the five estimation tasks (both politeness estimates and the stance estimate). The confidence-driven and non-adaptive approaches always achieve over 90\% coverage. In contrast, LLM-only coverage is always poor using GPT-3.5 (while using GPT-4o it was poor on four out of the five estimation tasks).

In summary, our insights regarding the gains of the confidence-driven approach are robust to the choices made in the LLM data collection.

\subsection{Sensitivity to Confidence Calibration}
\label{app:sensitivity}

A calibrated LLM will produce higher confidence scores when annotations are in \emph{agreement} with human annotations, compared to when annotations are in \emph{disagreement} with human annotations. However, calibration of confidence scores across tasks is not guaranteed.

To understand how the performance of \method is affected by the calibration of confidence scores, we conducted a robustness test, adding noise to confidence scores to simulate miscalibration. In particular, for illustration we consider the task of analyzing stance on global warming. We add a varying amount of normally distributed noise $\mathcal{N}(0,\sigma^{2})$ to the collected confidence scores $C_i$, and truncate the sum to $[0,1]$ to obtain a probability. 

We use a t-test to test the difference in calibration score means when LLM and human annotations agree, vs when LLM and human annotations disagree. If the t-statistic is large (equivalently, the corresponding p-value is small), that suggests that the two means differ significantly. As the random noise added to the confidence scores increases, the scores become less calibrated (Table \ref{tab:calibration}). We expect that our method performs worse in terms of $\neff$ when the confidence scores are miscalibrated, although coverage should be maintained. 

As predicted, as the amount of miscalibration in the confidence scores increases, the gain in the effective sample size decreases (Table \ref{tab:calibration}). The confidence-driven approach achieves the highest gain for the smallest amount of noise, though it always achieves a positive gain. This suggests that the approach is robust to poor confidence scores. Furthermore, \method achieves near 90\% coverage or higher in each setting, regardless of the amount of miscalibration.
Finally, we observe that \method achieves a higher gain than the non-adaptive approach regardless of the extent of miscalibration. This can be explained through the power tuning parameter $\lambda$; even when the confidence scores provide no signal, power tuning makes sure that LLM annotations are leveraged effectively.

\clearpage
\newpage

\begin{table*}[t]
\small
\begin{tabular}{p{0.1\textwidth}|p{0.1\textwidth}|p{0.7\textwidth}}
\toprule
\textbf{Setting} &  \textbf{Stage}    & \textbf{Prompt text} \\
 \midrule
\textbf{Politeness} & \textbf{Stage 1} & Is the following text polite? Output either A or B. Output a letter only.\\
&& A) Polite\\
&& B) Impolite\\
&&Text: <text>\\
&&Answer:\\
 \midrule
\textbf{Politeness} & \textbf{Stage 2} & How likely is it that the following text is <previously provided answer: polite or impolite>? \\
&& Output the probability only (a number between 0 and 1).\\
&& Text: <text>\\
&& Answer: \\
 \midrule
\textbf{Stance} & \textbf{Stage 1} &  A statement can agree, be neutral, or disagree with the statement: ``Climate change/global warming is a serious concern''. Classify the following statement into one of the three categories. Output either A, B, or C. Output a letter only.\\
&& A) Agree\\
&& B) Neutral\\
&& C) Disagree\\
&& Statement: <text> \\
&& Answer:\\
 \midrule
\textbf{Stance }& \textbf{Stage 2} & How likely is it that the following text <previously provided answer: agrees, neither agrees nor disagrees, or disagrees> with the statement: ``Climate change/global warming is a serious concern''?\\
&& Output the probability only (a number between 0 and 1).\\
&& Text: <text>\\
&& Probability: \\
 \midrule
\textbf{Bias} & \textbf{Stage 1}  & What is the political bias of the following article? Output either A,B, or C. Output a letter only.\\
& & A) Left \\
& & B) Center \\
& & C) Right \\
& & Article: <text>\\
& &  Answer:\\
\midrule
\textbf{Bias} & \textbf{Stage 2} & How likely is it that the following article has a <previously provided answer: left-leaning, centrist, or right-leaning> political bias? Output the probability only (a number between 0 and 1). \\
&&  Text: <text>\\
&&  Probability: \\
\bottomrule
\end{tabular}
\caption{\textbf{Complete prompt texts.} LLM annotation prompts across the three settings, for Stages 1 and 2.}
\label{tab:prompts}
\end{table*}

\begin{table*}[hp!]
\centering

\begin{tabular}{l|l|l}

\toprule
\multirow{2}{*}{\textbf{Method}}     & \multicolumn{2}{c}{\textbf{Metric}}                                                                  \\
                                     & \multicolumn{1}{c}{\textbf{Gain in eff. sample size}}     & \multicolumn{1}{c}{\textbf{Coverage}}    \\
\midrule
\midrule
\multicolumn{3}{l}{\textbf{default experiment, cf. Figure~\ref{fig:results}}} \\
\multicolumn{3}{l}{\textbf{(model: GPT-4o, prompting: zero-shot, annotation task: binary classification)}} \\
\multicolumn{3}{l}{{Cohen inter-rater agreement $\kappa_{\mathrm{politeness}}=0.39$}}                                                \\
\midrule
LLM only                             & ---                                                       & \hlsecondarytab{71\%}                                     \\
human + LLM (non-adaptive)           & \hlsecondarytab{-9.85\%}                                                   & \hlprimarytab{93\%}                                     \\
confidence-driven                    & \hlprimarytab{\textbf{248\%}}                                                     & \hlprimarytab{99\%}                                     \\
\midrule
\midrule
\multicolumn{3}{l}{\textbf{alternative model: GPT-4o-mini}}                                                                                 \\
\multicolumn{3}{l}{{Cohen inter-rater agreement $\kappa_{\mathrm{politeness}}=0.56$}}                                                \\
\midrule
LLM only                             & ---                                                       & \hlprimarytab{90\%}                                     \\
human + LLM (non-adaptive)           & \hlprimarytab{24.15\%}                                                   & \hlprimarytab{94\%}                                     \\
confidence-driven                    & \hlprimarytab{\textbf{265.69\%}}                                                  & \hlprimarytab{100\%}                                    \\
\midrule
\midrule
\multicolumn{3}{l}{\textbf{alternative prompting: few-shot (10 examples)}}                                                                  \\
\multicolumn{3}{l}{{Cohen inter-rater agreement $\kappa_{\mathrm{politeness}}=0.45$}}                                                \\
\midrule
LLM only                             & ---                                                       & \hlsecondarytab{82\%}                                     \\
human + LLM (non-adaptive)           & \hlprimarytab{3.1\%}                                                     & \hlprimarytab{91\%}                                     \\
confidence-driven                    & \hlprimarytab{\textbf{251.25\%}}                                                  & \hlprimarytab{100\%}                                    \\
\midrule
\midrule
\multicolumn{3}{l}{\textbf{alternative annotation task: Likert scale (7-point)}}                                                            \\
\multicolumn{3}{l}{{Cohen inter-rater agreement $\kappa_{\mathrm{politeness}}=0.21$}}                                                \\
\midrule
LLM only                             & ---                                                       & \hlsecondarytab{26\%}                                     \\
human + LLM (non-adaptive)           & \hlsecondarytab{-27.45\%}                                                  & \hlprimarytab{92\%}                                     \\
confidence-driven                    & \hlprimarytab{\textbf{239.92\%}}                                                  & \hlprimarytab{99\%}        \\
\bottomrule
\end{tabular}
\caption{\textbf{Sensitivity to the LLM data collection method.} Gain in effective sample size and coverage for the LLM only, human + LLM (non-adaptive), and confidence-driven approaches, across varying data collection approaches. Results are presented for the task of analyzing impact of hedging on perceived politeness, $\nhuman=1100$, estimated over 100 trials. The confidence-driven approach always achieves a large ~\hlprimarytab{positive gain}. For each LLM data collection method, the confidence-driven approach achieves a higher gain in effective sample size than the non-adaptive approach (marked in \textbf{bold}); it also achieves \hlprimarytab{near 90\% coverage} or higher in each setting. In contrast, LLM-only coverage is often \hlsecondarytab{poor}. Gain in effective sample size is not estimated for the LLM-only approach as it does not leverage human annotations.}
\label{tab:robustness}
\end{table*}

\begin{table*}[t!]
\small
\centering
\begin{tabular}{l|l|rrrr}
\toprule
\multirow{2}{*}{\textbf{Estimation task}}                & \multirow{2}{*}{\textbf{Metric}}                   &     \multicolumn{3}{c}{\textbf{Method}}   \\
&  & confidence-driven & human + LLM (non-adaptive)  & LLM only \\
\midrule

\multirow{2}{*}{\textbf{\shortstack[l]{Politeness devices \\ (hedge)}}} & \textbf{Gain in eff. sample size } & (\hlprimarytab{\textbf{31.51}} $\pm$ 7.81)\%  & (\hlsecondarytab{-12.23} $\pm$ 9.18)\%  & --- \\
& \textbf{Coverage} & \hlprimarytab{
92\%} & \hlprimarytab
{92\%} & \hlsecondarytab{39\%} \\
\midrule

\multirow{2}{*}{\textbf{\shortstack[l]{Politeness devices \\ (1st person plural)}}} & \textbf{Gain in eff. sample size } & (\hlprimarytab{\textbf{321.00}} $\pm$ 19.01)\%  & (\hlsecondarytab{-5.77}$\pm$ 30.83)\%  & --- \\
& \textbf{Coverage} & \hlprimarytab
{97\%} & \hlprimarytab
{92\%} & \hlsecondarytab{67\%} \\
\midrule
                                 
\multirow{2}{*}{\textbf{\shortstack[l]{Stance on \\ global warming}}} & \textbf{Gain in eff. sample size} & (\hlprimarytab{\textbf{82.68}} $\pm$ 9.17)\% & (\hlsecondarytab{-11.23} $\pm$ 17.15)\%  &  --- \\
& \textbf{Coverage} &  \hlprimarytab
{96\%}  & \hlprimarytab
{94\%} & \hlsecondarytab{0\%}   \\
\midrule
\multirow{2}{*}{\textbf{\shortstack[l]{Political bias \\ (left-leaning)}}} & \textbf{Gain in eff. sample size} & (\hlprimarytab{\textbf{17.08}} $\pm$ 6.30)\%  & (\hlprimarytab{3.83} $\pm$ 4.86)\%  &  ---                 \\
 & \textbf{Coverage} & \hlprimarytab{93\%} & \hlprimarytab
{98\%} & \hlsecondarytab{18\%}  \\
 \midrule
 \multirow{2}{*}{\textbf{\shortstack[l]{Political bias \\ (right-leaning)}}} & \textbf{Gain in eff. sample size} & (\hlprimarytab{\textbf{23.10}} $\pm$ 7.50)\%  & (\hlprimarytab{11.09} $\pm$ 5.73)\%  &  ---                 \\
 & \textbf{Coverage} & \hlprimarytab
{95\%} & \hlprimarytab
{94\%} & \hlsecondarytab{11\%}  \\
\bottomrule
\end{tabular}
\caption{\textbf{Results summary (GPT-3.5).} Gain in effective sample size and coverage across the five estimation tasks for $\nhuman=500$, estimated over 100 trials.  In each task, the confidence-driven approach achieves a higher gain in effective sample size (\textbf{bolded}) than the non-adaptive approach. Confidence-driven approach always achieves a~\hlprimarytab{positive gain}, while the non-adaptive approach sometimes achieves a~\hlsecondarytab{negative gain}. Confidence-driven and non-adaptive approaches achieve \hlprimarytab{near 90\% coverage}, or higher. In contrast, LLM-only coverage is \hlsecondarytab{poor}. Gain in effective sample size is not estimated for the LLM-only approach as it does not leverage human annotations. Errors show a standard deviation over 100 trials.}
\label{tab:3.5}
\end{table*}

\begin{table*}[t]
\centering
\small
\begin{tabular}{l|l|r|r}
\toprule
\multirow{2}{*}{\textbf{Method}}              & \multirow{2}{*}{\textbf{Confidence score calibration t-test}}              & \multicolumn{2}{c}{\textbf{Metric}}          \\
                                  &   & \textbf{Gain in eff. sample size} & \textbf{Coverage} \\
\midrule
\multicolumn{4}{l}{\textbf{$\nhuman=500$}}\\
\midrule
\textbf{LLM only}                             & --- &  ---                      & \hlsecondarytab{0\%}      \\
\textbf{human + LLM (non-adaptive)}           & --- & \hlsecondarytab{-3.46\%}                  & \hlprimarytab{100\%}    \\
\textbf{confidence-driven ($\sigma^2 = 0$) }  & 
$t = 9.08$, $p = 2.19\times 10^{-19}$ & \hlprimarytab{\textbf{43.48\% }}                 & \hlprimarytab{94\%}     \\
\textbf{confidence-driven ($\sigma^2 = 0.2$)} & $t = 5.53$, $p = 3.65 \times 10^{-8}$ & \hlprimarytab{40.70\%}                  & \hlprimarytab{95\%}    \\
\textbf{confidence-driven ($\sigma^2 = 0.4$)} & $t =3.30$, $p = 0.000994$ & \hlprimarytab{42.82\%}                  & \hlprimarytab{94\%}     \\
\textbf{confidence-driven ($\sigma^2 = 0.6$)} & $t =2.34$, $p = 0.0192$ & \hlprimarytab{39.57\%}                  & \hlprimarytab{93\%}    \\
\textbf{confidence-driven ($\sigma^2 = 0.8$)} & $t =1.88$, $p = 0.0606$ & \hlprimarytab{40.87\%}                  & \hlprimarytab{94\%}  \\  
\midrule
\multicolumn{4}{l}{\textbf{$\nhuman=1150$}}\\
\midrule
\textbf{LLM only}                             & --- &  ---                      & \hlsecondarytab{0\%}      \\
\textbf{human + LLM (non-adaptive)}           & --- & \hlprimarytab{17.02\%}                  & \hlprimarytab{100\%}    \\
\textbf{confidence-driven ($\sigma^2 = 0$) }  & 
$t = 9.08$, $p = 2.19\times 10^{-19}$ & \hlprimarytab{\textbf{28.14\% }}                 & \hlprimarytab{99\%}     \\
\textbf{confidence-driven ($\sigma^2 = 0.2$)} & $t = 5.29$, $p = 1.36 \times 10^{-7}$ & \hlprimarytab{25.30\%}                  & \hlprimarytab{100\%}    \\
\textbf{confidence-driven ($\sigma^2 = 0.4$)} & $t =3.22$, $p = 0.00130$ & \hlprimarytab{25.84\%}                  & \hlprimarytab{97\%}     \\
\textbf{confidence-driven ($\sigma^2 = 0.6$)} & $t =2.23$, $p = 0.0257$ & \hlprimarytab{26.81\%}                  & \hlprimarytab{100\%}    \\
\textbf{confidence-driven ($\sigma^2 = 0.8$)} & $t =1.83$, $p = 0.0831$ & \hlprimarytab{25.71\%}                  & \hlprimarytab{99\%}  \\  
\bottomrule
\end{tabular}
\caption{\textbf{Sensitivity to confidence score calibration.} Gain in effective sample size and coverage for the LLM only, human + LLM (non-adaptive), and confidence-driven approaches, given varying amounts of miscalibration in confidence scores ($\sigma^2$). Results are presented for the task of analyzing stance on global warming, estimated over 100 trials. The t-test tests for the difference in calibration score means when LLM and human annotations agree, vs when LLM and human annotations disagree (larger $t$ means difference is more significant). The confidence-driven approach achieves the largest gain for the smallest amount of noise (\textbf{bolded}), and it always achieves a~\hlprimarytab{positive gain}. For each amount of miscalibration, the confidence-driven approach achieves a higher gain in effective sample size than the non-adaptive approach; it also achieves \hlprimarytab{near 90\% coverage} or higher in each setting. In contrast, LLM-only coverage is \hlsecondarytab{poor}. Gain in effective sample size is not estimated for the LLM-only approach as it does not leverage human annotations.  }
\label{tab:calibration}
\end{table*}

\end{document}